\documentclass[default,iicol]{sn-jnl}

\usepackage[table,xcdraw]{xcolor}

\usepackage{graphicx}%
\usepackage{multirow}%
\usepackage{amsmath,amssymb,amsfonts}%
\usepackage{amsthm}%
\usepackage{mathrsfs}%
\usepackage[title]{appendix}%
\usepackage{xcolor}%
\usepackage{textcomp}%
\usepackage{manyfoot}%
\usepackage{booktabs}%
\usepackage{algorithm}%
\usepackage{algorithmicx}%
\usepackage{listings}%
\usepackage[capitalise]{cleveref }
\usepackage{algpseudocode}
\usepackage{textcomp}
\usepackage{url} 
\usepackage{makecell} 
\usepackage{soul}
\usepackage{booktabs}
\usepackage{multirow}
\usepackage{siunitx}
\usepackage{dblfloatfix}
\usepackage{float} 
\usepackage{geometry}
\theoremstyle{./bst/thmstyleone}%
\theoremstyle{thmstyletwo}%

\theoremstyle{thmstylethree}%

\algnewcommand\algorithmicforeach{\textbf{for each}}
\algdef{S}[FOR]{ForEach}[1]{\algorithmicforeach\ #1\ \algorithmicdo}

\begin{document}

\title[Article Title]{A Novel Voronoi-based Convolutional Neural Network Framework for Pushing Person Detection in Crowd Videos}


\author*[1,2,3]{\fnm{Ahmed} \sur{Alia}}\email{a.alia@fz-juelich.de}

\author[4]{\fnm{Mohammed} \sur{ Maree}}\email{mohammed.maree@aaup.edu}
 
\author[1]{\fnm{Mohcine} \sur{ Chraibi}}\email{m.chraibi@fz-juelich.de}

\author[1,2]{\fnm{Armin} \sur{ Seyfried}}\email{a.seyfried@fz-juelich.de}

\affil[1]{\orgdiv{Institute for Advanced Simulation}, \orgname{ Forschungszentrum Jülich}, \orgaddress{ \city{Jülich}, \postcode{52425},  \country{Germany}}}

\affil[2]{\orgdiv{Department of Computer Simulation for Fire Protection and Pedestrian Traffic, Faculty of Architecture and Civil Engineering}, \orgname{ University of Wuppertal}, \orgaddress{ \city{Wuppertal}, \postcode{42285},  \country{Germany}}}

\affil[3]{\orgdiv{Department of Information Technology}, \orgname{An-Najah National University}, \orgaddress{ \city{Nablus},   \country{Palestine}}}

\affil[4]{\orgdiv{Department of Information Technology}, \orgname{Arab American University}, \orgaddress{ \city{Jenin},   \country{Palestine}}}


\abstract{
Analyzing the microscopic dynamics of pushing behavior within crowds can offer valuable insights into crowd patterns and interactions. By identifying instances of pushing in crowd videos, a deeper understanding of when, where, and why such behavior occurs can be achieved. This knowledge is crucial to creating more effective crowd management strategies, optimizing crowd flow, and enhancing overall crowd experiences. However, manually identifying pushing behavior at the microscopic level is challenging, and the existing automatic approaches cannot detect such microscopic behavior. Thus, this article introduces a novel automatic framework for identifying pushing in videos of crowds on a microscopic level. The framework comprises two main components: i) Feature extraction and ii) Video labeling. In the feature extraction component, a new Voronoi-based method is developed for determining the local regions associated with each person in the input video. Subsequently, these regions are fed into EfficientNetV1B0 Convolutional Neural Network to extract the deep features of each person over time. In the second component, a combination of a fully connected layer with a Sigmoid activation function is employed to analyze these deep features and annotate the individuals involved in pushing within the video. The framework is trained and evaluated on a new dataset created using six real-world experiments, including their corresponding ground truths. The experimental findings indicate that the suggested framework outperforms seven baseline methods that are employed for comparative analysis purposes.}




\maketitle

\section{Introduction}\label{sec1}

With the rapid development of urbanization, the dense crowd has become widespread in various locations, such as religious sites, train stations, concerts, stadiums, malls, and famous tourist attractions. 
In such highly dense crowds, pushing behavior can easily arise. Such behavior could increase the crowd's density, potentially posing a threat not only to people's comfort but also to their safety~\cite{feldmann2023forward, li2021experimental, sieben2023inside, adrian2020crowds, wang2018study}. 
People in crowds start pushing for different reasons. It could be for saving their lives from fire~\cite{keating1982myth, sime1983affiliative, li2022effect} or other hazards, catching a bargain on sail or simply accessing an overcrowded subway train~\cite{goyal2020analysis, CroMA}, and gaining access to a venue~\cite{johnson1987panic, helbing2012crowd, adrian2020crowding, sieben2023inside, adrian2020crowds}.
Understanding the microscopic dynamics of pushing plays a pivotal role in effective crowd management, helping safeguard the crowd from tragedies and promoting overall well-being~\cite{wang2020experimental, feldmann2023forward}. This has led to several studies aiming to comprehend pushing dynamics, especially in crowded event entrances~\cite{usten2022pushing, alia2022hybrid, alia2023cloud, CrowdDNA, BaSiGo, metivet2018push}. Lügering et al.~\cite{usten2022pushing} defined pushing as a behavior that pedestrians use to reach a target (like accessing an event) faster. This behavior involves pushing others using arms, shoulders, elbows, or the upper body, as well as utilizing gaps among neighboring people to navigate forward quicker. 
The study~\cite{usten2022pushing} has introduced a manual rating to understand pushing dynamics at the microscopic level. The method relies on two trained psychologists to classify pedestrians' behaviors over time in a video of crowds into pushing or non-pushing categories, helping to know when, where, and why pushing behavior occurs. 
However, this manual method is time-consuming, tedious and prone to errors in some scenarios.
Additionally, it requires trained observers, which may not always be feasible. Consequently, an increasing demand is for an automatic approach to identify pushing at the microscopic level within crowd videos.
Detecting pushing behavior automatically is a demanding task that falls within the realm of computer vision. This challenge arises from several factors, such as dense crowds gathering at event entrances, the varied manifestations of pushing behavior, and the significant resemblance and overlap between pushing and non-pushing actions.

Recently, machine learning algorithms, particularly Convolutional Neural Network (CNN) architectures, have shown remarkable success in various computer vision tasks, including face recognition~\cite{budiman2023student}, object
detection~\cite{lu2023cnn}, and abnormal behavior detection~\cite{direkoglu2020abnormal}. 
One of the key reasons for this success is that CNN can learn the relevant features~\cite{alia2017feature, alia2021enhanced, alia2016hybrid} automatically from data without human supervision~\cite{gan2022spatiotemporal,gan2021automated}.
As a result of CNN's success in abnormal behavior detection, which is closely related to pushing detection, some studies have started to automate pushing detection using CNN models~\cite{alia2022hybrid,alia2022fast, alia2023cloud}.
For instance, Alia et al.~\cite{alia2022hybrid, alia_ahmed_2023_6433908} introduced a deep learning framework that leverages deep optical flow and CNN models for pushing patch detection in video recordings. Another study~\cite{alia2022fast} introduced a fast hybrid deep neural network model based on GPU to enhance the speed of video analysis and pushing patch identification.
Similarly, the authors of~\cite{alia2023cloudpre, alia2023cloud, alia_ahmed_2023_7570208} developed an intelligent framework that combines deep learning algorithms, a cloud environment, and live camera stream technology to annotate the pushing patches in real-time from crowds accurately.
Yet, the current automatic methods focus on identifying pushing behavior at the level of regions (macroscopic level) rather than at the level of individuals (microscopic level), where each region can contain a group of persons.  
In other words, the automatic approaches reported in the literature can not detect pushing at the microscopic level, limiting their contributions to help comprehend pushing dynamics in crowds. For example, they cannot accurately determine the relationship between the number of individuals involved in pushing behavior and the onset of critical situations, thereby hindering a precise understanding of when a situation may escalate to a critical level.

To overcome the limitations of the aforementioned methods, this article introduces a novel Voronoi-based CNN framework for automatically identifying instances of microscopic pushing behavior from crowd video recordings. 
The proposed framework comprises two components: feature extraction and labeling. The first component utilizes a novel Voronoi-based EfficientNetV1B0 CNN architecture for feature extraction. The Voronoi~\cite{green1978computing}-based method is used to identify the local region of each person over time, and then the EfficientNetV1B0 model~\cite{tan2019efficientnet} extracts deep features from these regions. In this article, the local region is defined as the zone focusing only on a single person (target person), including his surrounding spaces and physical interactions with his direct neighbors. This region is crucial in guiding the proposed framework to focus on microscopic behavior. 
On the other hand, the second component employs a fully connected layer with a Sigmoid activation function to analyze the deep features and detect the pushing persons. 
The framework (CNN and fully connected layer) is trained from scratch on a dataset of labeled local regions generated from six real-world video experiments with their ground truths~\cite{crowdqueue}.

The main contributions of this work are summarized as
follows:
\begin{enumerate}
\item To the best of our knowledge, this article presents the first framework for automatically identifying pushing at the individual level in videos of human crowds.

\item This article introduces a novel feature extraction method for characterizing microscopic behavior in videos of crowds, particularly pushing behavior.

\item The article creates a fresh dataset derived from local regions and includes data from six real-world experiments, each paired with corresponding ground truths. This dataset represents a valuable resource for future research in this domain.

\end{enumerate}

The remainder of this article is organized as follows.
Section~\ref{sec:relatedwork} reviews some automatic approaches to abnormal behavior detection in videos of crowds.
The architecture of the proposed framework is introduced in ~\cref{sec:proposedframework}.
\cref{trainingandevaluatingtheframework} presents the processes of training and evaluating the framework. \cref{sec:evaluationandresults} discusses experimental results and comparisons.  Finally, the conclusion and future work are summarized in~\cref{sec:conclusionandfuturework}.

\section{Related Work}
\label{sec:relatedwork}
This section begins by providing an overview of CNN-based approaches for automatic video analysis and detecting abnormal behavior in crowds. 
It then discusses the methods for automatically detecting pushing patches in crowd videos.

\subsection{CNN-based Abnormal Behavior Detection }
Typically, behavior is considered abnormal when seen as unusual under specific contexts.
This implies that the definition of abnormal behavior depends on the situation~\cite{tay2019robust}.
To illustrate, running inside a bank might be considered abnormal behavior, whereas running at a traffic light could be viewed as normal~\cite{duman2019anomaly}.
Several behaviors have been addressed automatically in abnormal behavior detection applications in crowds, including walking in the wrong direction~\cite{roshtkhari2013line}, running away~\cite{singh2019crowd}, sudden people grouping or dispersing~\cite{george2018crowd},  human falls~\cite{santos2019accelerometer}, suspicious behavior, violent acts~\cite{mehmood2021lightanomalynet}, abnormal crowds~\cite{zhang2018energy}, hitting,  and kicking~\cite{kooij2016multi}.

Tay et al.~\cite{tay2019robust} developed a CNN-based method for identifying abnormal activities from videos. The researchers specifically designed and trained a customized CNN to extract features and label samples, utilizing a dataset comprising both normal and abnormal samples.
In another study, Alafif et al.~\cite{alafif2023hybrid} introduced two approaches for detecting abnormal behaviors in crowd videos, varying in scale from small to large.
For detecting anomaly behaviors in a small-scale crowd at the object level, the first method utilizes a hybrid approach that combines a pre-trained CNN model with a random forest classifier.
On the other hand, the second method employs a two-step approach to identify abnormal behaviors in a large-scale crowd. Initially, a pre-trained model is used as the first classifier to identify frames containing abnormal behaviors. Subsequently, the second classifier, specifically You Only Look Once (version 2), is utilized to analyze the identified frames and detect abnormal behaviors exhibited by individuals. 
Nevertheless, constructing an accurate CNN classifier requires a substantial training dataset, often unavailable for many human behaviors.

To address the limited availability of large datasets containing both normal and abnormal behaviors, some researchers have employed one-class classifiers using datasets that exclusively consist of normal behaviors. 
Creating or acquiring a dataset containing only normal behavior is comparatively easier than obtaining a dataset that includes both normal and abnormal behaviors.~\cite {sabokrou2018deep,xu2019efficient}.
The fundamental concept behind the one-class classifier is to learn exclusively from normal behaviors, thereby establishing a class boundary between normal and undefined (abnormal) classes.
For example,
Sabokrou~et~al.~\cite {sabokrou2018deep}  utilized a pre-trained CNN to extract motion and appearance information from crowded scenes. They then employed a one-class Gaussian distribution to build the classifier, utilizing datasets of normal behavior.
Similarly, in~\cite{xu2019efficient,smeureanu2017deep}, the authors constructed one-class classifiers by leveraging a dataset composed exclusively of normal samples.
In~\cite{xu2019efficient}, Xu et al. employed a convolutional variational autoencoder to extract features, followed by the use of multiple Gaussian models to detect abnormal behavior. Meanwhile, in~\cite{smeureanu2017deep}, a pre-trained CNN model was employed for feature extraction, while one-class support vector machines were utilized for detecting abnormal behavior.
Another study by Ilyas et al. \cite{ilyas2021hybrid} 
conducted a separate study where they utilized a pre-trained CNN along with a gradient sum of the frame difference to extract meaningful features. Subsequently, they trained three support vector machines on normal behavior data to identify abnormal behaviors.
In general, one-class classifiers are frequently employed when the target behavior class or abnormal behavior is rare or lacks a clear definition~\cite{khan2014one}. However, pushing behavior is well-defined and not rare, particularly in high-density and competitive situations. Furthermore, this type of classifier considers new normal behavior as abnormal.

In order to address the limitations of CNN-based and one-class classifier approaches, multiple studies have explored the combination of multi-class CNNs with one or more handcrafted feature descriptors~\cite {ilyas2021hybrid,direkoglu2020abnormal}.
In these hybrid approaches, the descriptors are employed to extract valuable information from the data. Subsequently, CNN learns and identifies relevant features and classifications based on the extracted information.
For instance, Duman et al.~\cite{duman2019anomaly} employed the classical Farneb\"{a}ck optical flow method~\cite{farneback2003two} and CNN to identify abnormal behavior.
They used Farneb\"{a}ck and CNN to estimate direction and speed information and then applied a convolutional long short-term memory network to build the classifier.
Hu et al.~\cite{hu2020design} employed a combination of the histogram of gradient and CNN for feature extraction, while a least-squares support vector was used for classification.
Direkoglu~\cite{direkoglu2020abnormal} utilized the Lucas-Kanade optical flow method and CNN to extract relevant features and identify ``escape and panic behaviors''.
Almazroey et al.~\cite{almazroey2020abnormal}  used Lucas-Kanade optical flow, a pre-trained CNN, and feature selection methods (specifically neighborhood component analysis) to extract relevant features. These extracted features were then used to train a support vector machine classifier.
In another study~\cite{zhou2016spatial}, Zhou et al. introduced an approach based on CNN for detecting and localizing anomalous activities. Their approach involved integrating optical flow with a CNN for feature extraction and utilizing a CNN for the classification task.

Hybrid-based approaches could be more suitable for automatically detecting pushing behavior due to the limited availability of labeled pushing data. 
Nevertheless, most of the reviewed hybrid-based approaches for abnormal behavior detection may be inefficient for detecting pushing since 1) The descriptors used in these approaches can only extract limited essential data from high-density crowds to represent pushing behavior.
2) Some CNN architectures commonly utilized in these approaches may not be effective in dealing with the increased variations within pushing behavior (intra-class variance) and the substantial resemblance between pushing and non-pushing behaviors (high inter-class similarity), which can potentially result in misclassification.

\begin{figure*}[h!]
   \centering
   \includegraphics[width=1\textwidth]{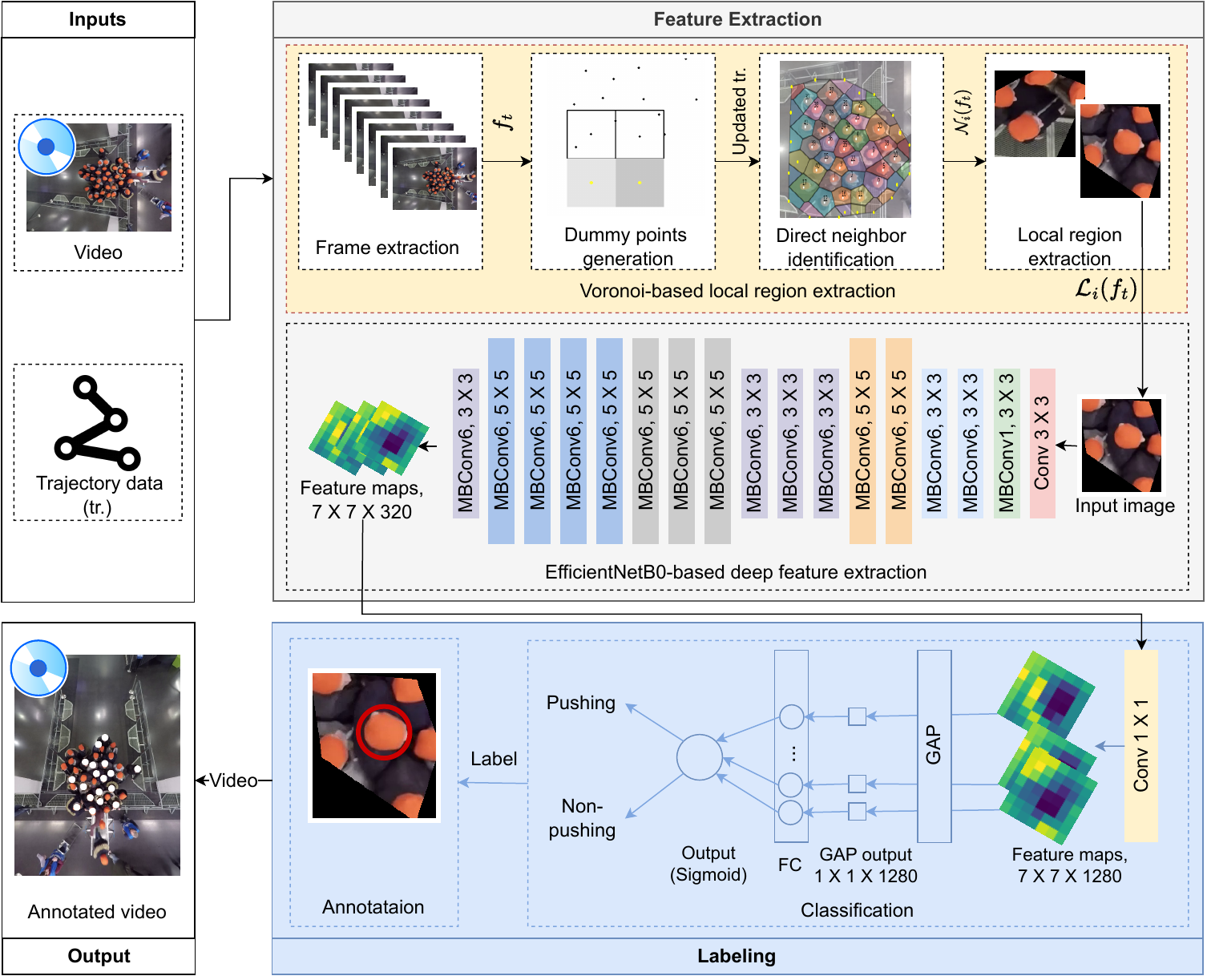}
  \caption{The architecture of the proposed framework.  In \(f_t\), \(f\) signifies an extracted frame, while \(t\) indicates its timestamp in seconds, counted from the beginning of the input video (with \(t\) taking values like \(1, 2, 3, \dots\)).
  For a target person \(i\) at \(f_t\), \(\mathcal{L}_i(f_t)\) denotes the local region, while \(\mathcal{N}_i(f_t)\) represents the direct neighbors.
  FC stands for fully connected layer, while GAP refers to global average pooling.}
  \label{fig:framwork}
\end{figure*}

\subsection{CNN-based Pushing Behavior Detection}

In more recent times, a few approaches that merge effective descriptors with robust CNN architectures have been developed for detecting pushing regions in crowds.
For example,  Alia et al.~\cite{alia2022hybrid} introduced a hybrid deep learning and visualization framework to aid researchers in automatically detecting pushing behavior in videos.
The framework combines deep optical flow and visualization methods to extract the visual motion information from the input video. 
This information is then analyzed using an EfficientNetV1B0-based CNN and false reduction algorithms to identify and label pushing patches in the video.
The framework has a drawback in terms of speed, as the motion extraction process is based on a CPU-based optical flow method, which is slow.
Another study~\cite{alia2022fast} presented a fast hybrid deep neural network model that labels pushing patches in short videos lasting only two seconds.
The model is based on an EfficientNetB1-based CNN and GPU-based deep optical flow. 

To support the early detection of pushing patches within crowds,
the study~\cite{alia2023cloud} presented a cloud-based deep learning system. The primary goal of such a system is to offer organizers and security teams timely and valuable information that can enable early intervention and mitigate hazardous situations.
The proposed system relies mainly on a fast and accurate pre-trained deep optical flow, an adapted version of the EfficientNetV2B0-based CNN, a cloud environment and live stream technology.
Simultaneously, the optical flow model extracts motion characteristics of the crowd in the live video stream, and the classifier analyzes the motion to label pushing patches directly on the stream.
Moreover, the system stores the annotated data in the cloud storage, which is crucial to assist planners and organizers in evaluating their events and enhancing their future plans.

To the best of our knowledge, current pushing detection approaches in the literature primarily focus on identifying pushing at the patch level rather than at the individual level.
However, identifying the individuals involved in pushing would be more helpful for understanding the pushing dynamics. 
Hence, this article introduces a new framework for detecting pushing individuals in videos of crowds. The following section provides a detailed discussion of the framework.

\section{Proposed Framework Architecture}
\label{sec:proposedframework}

\begin{figure*}[h]
   \centering
   \includegraphics[width=1\textwidth]{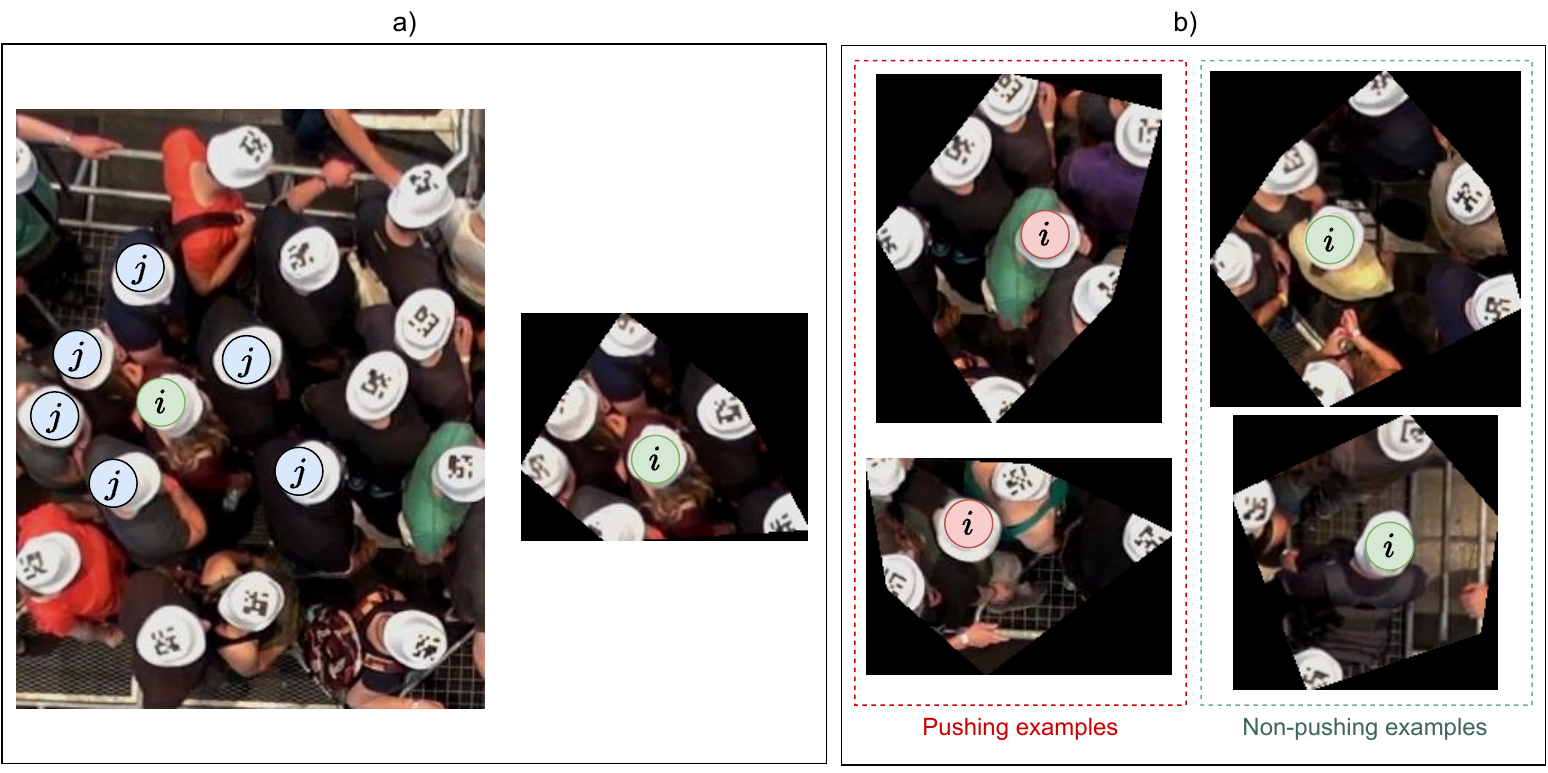}
  \caption{An illustration of direct neighbors (a) and examples of local regions (b).  The red circles represent individuals engaged in pushing, while the green circles represent individuals not involved in pushing. Direct neighbors $j$ of a person $i$ are indicated with blue circles.}
  \label{fig:LR-crowdexamples}
\end{figure*}

This section describes the proposed framework for automatic pushing person detection in videos of crowds. As depicted in~\cref{fig:framwork}, there are two main components: feature extraction and labeling. The first component extracts the deep features from each individual’s behavior. In contrast,  the second component analyzes the extracted deep features and annotates the pushing persons within the input video. The following sections will discuss both components in more detail.

\subsection{Feature Extraction Component}

This component aims to extract deep features from each individual's behavior, which can be used to classify pedestrians as pushing or non-pushing. To accomplish this, the component consists of two modules: Voronoi-based local region extraction and EfficientNetV1B0-based deep feature extraction. The first module selects a frame every second from the input video and identifies the local region of each person within those extracted frames. Subsequently, the second module extracts deep features from each local region and feeds them to the next component for pedestrian labeling. Before diving into these modules, let us define the local region term at one frame. 

A frame \(f_t\) is captured every second from the input video. Here, \(t\) represents the timestamp, in seconds, since the start of the video and can range from 1 to \(T\), where \(T\) is the total duration of the video in seconds.
We can analyze individual pedestrians within each of these frames, such as \(f_t\). For instance, consider a pedestrian \(i\) positioned at \(\langle x, y\rangle_i\). 
Let \(\mathcal{N}_i\) denote the set of pedestrians whose Voronoi cells are adjacent to that of pedestrian \(i\). 
Specifically, pedestrian \(j\) belongs to \(\mathcal{N}_i\) if and only if their Voronoi cells share a boundary. 
The local region for pedestrian \(i\) at \(f_t\), \(\mathcal{L}_i\), forms a two-dimensional closed polygon, defined by the positions of all pedestrians in \(\mathcal{N}_i\). 
As illustrations, \cref{fig:LR-crowdexamples}a provides examples of both \(\mathcal{N}_i\) (left image) and \(\mathcal{L}_i\) (right image).

The region \(\mathcal{L}_i\) encapsulates the crowd dynamics around individual \(i\), reflecting potential interactions between \(i\) and its neighbors \(\mathcal{N}_i\). 
Notably, the characteristics around a pushing individual might diverge from those around a non-pushing one, a distinction pivotal for highlighting pushing behaviors. \cref{fig:LR-crowdexamples}b showcases examples of such \(\mathcal{L}_i\) regions for pushing and non-pushing individuals. 
The following section introduces a novel method for extracting \(\mathcal{L}_i\).

\subsubsection{Voronoi-based Local Region Extraction}

\begin{figure*}
   \centering
   \includegraphics[width=1\textwidth]{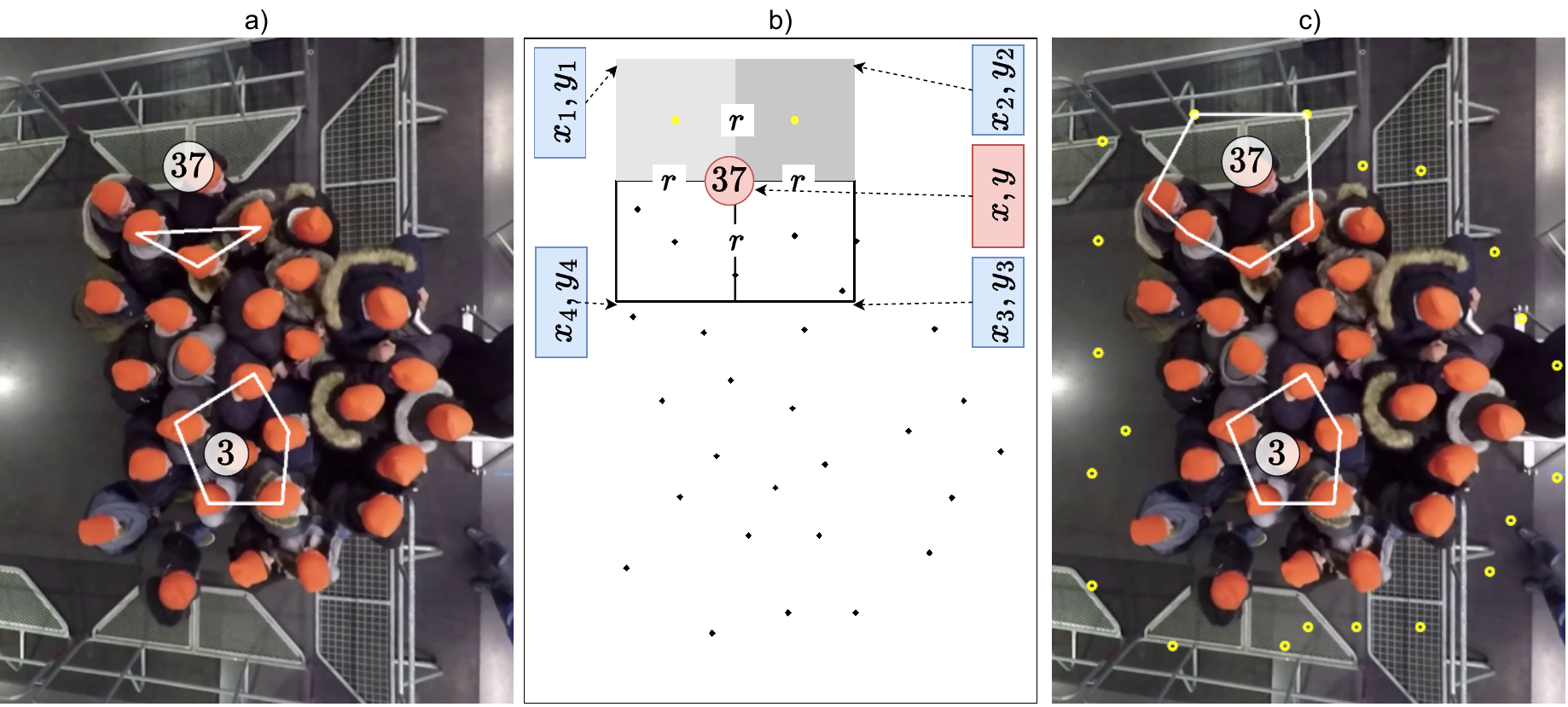}
  \caption{An illustration of the effect of dummy points on creating the local regions, as well as a sketch of the dummy points generation technique. a) $\mathcal{L}_{37}$ and $\mathcal{L}_3$ without dummy points. b)  a sketch of the dummy points generation technique.  c) $\mathcal{L}_{37}$ and $\mathcal{L}_3$ with dummy points.  The white polygon represents the border of the local regions. Yellow small circles refer to the generated dummy points, while black points in b denote the positions of pedestrians. $r$ is the dimension of each square.  }
  \label{fig:dummyxamples}
\end{figure*}

\begin{figure*}[h!]
   \centering
   \includegraphics[width=0.8\textwidth]{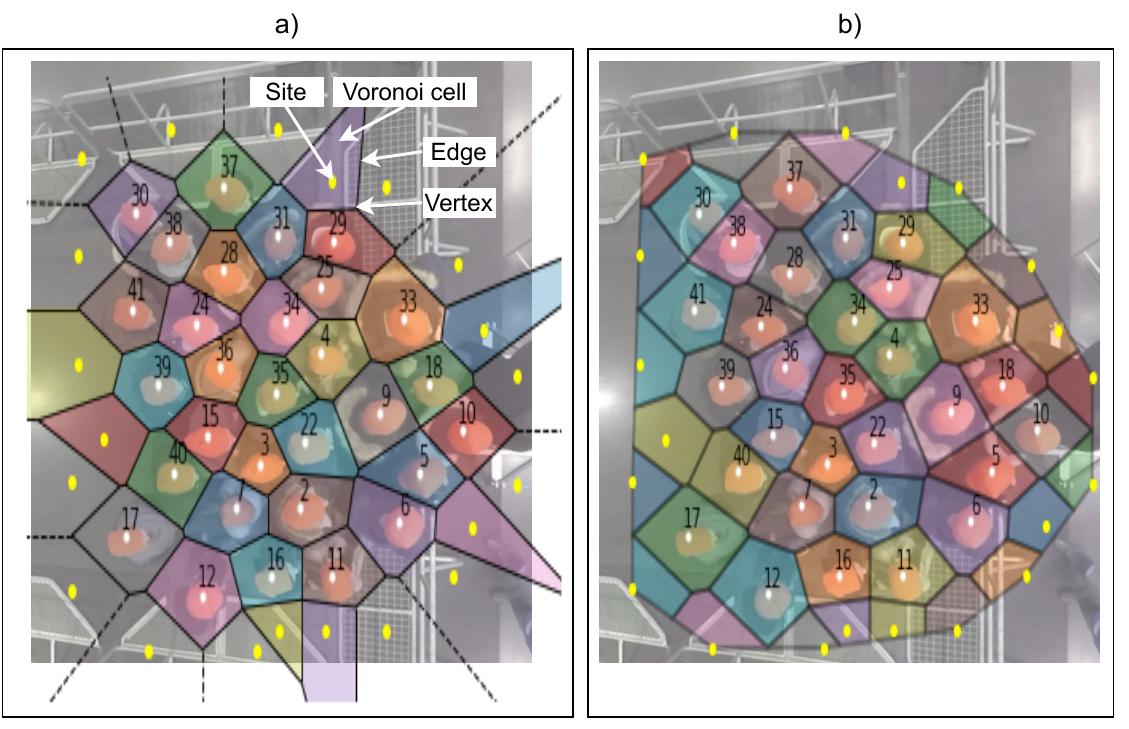}
  \caption{a) Example of a simple Voronoi decomposition. b) Example of bounded Voronoi decomposition. Both are constructed using 30 pedestrian points and 21 dummy points.}
  \label{fig:voronoi}
\end{figure*} 

This section presents a novel method for extracting the local regions of pedestrians from the input video over time.  The technique consists of several steps: frame extraction,  dummy points generation, direct neighbor identification, and local region extraction.  

\begin{algorithm*}[b]
\caption{Pseudo code for generating dummy points.}\label{alg:dummypoints}  
 \textbf{Inputs:} \\
 \hspace*{\algorithmicindent} \textbf{$tr$}: a file of pedestrian trajectory data over frames,\\ 
 \hspace*{\algorithmicindent}\hspace*{\algorithmicindent} where each record represents [person Id, frame order, x-coordinate, y-coordinate]  \\
 \hspace*{\algorithmicindent} \textbf{$fps$}: the frame rate of the input video, measured in frames per second.   \\
 \hspace*{\algorithmicindent} \textbf{$r$}: the dimension of each  square region.  \\ 
 \textbf{Outputs:} \\
 \hspace*{\algorithmicindent} \textbf{$tr\_dummy$}: a file of pedestrian trajectory data (over seconds) with dummy points.    

\begin{algorithmic}[1]

\State $file \gets $ open($tr$)
\State $file\_dummy \gets $ open($tr\_dummy$)
\While{not EOF($file$)}
    \State $rec \gets$ read($file$) 
    \If{$rec[1] \,\%\, fps \, = \,0$}
        \State{write(rec) to \,$tr\_dummy$}
    \EndIf
\EndWhile

\State $regions \gets [\,[\,]\,]$
\While{not EOF($file\_dummy$)}
    \State $rec \gets$ read($file\_dummy$) 
     
    \State $x \gets rec[2]$
    \State $y \gets rec[3]$
    \State append ([$x-r$, $y+r$]) to regions
    \State append ([$x+r$, $y+r$]) to regions
    \State append ([$x+r$, $y-r$]) to regions
    \State append ([$x-r$, $y-r$]) to regions
        
    \While{$corner $ in $  regions$}
            \If{empty(area([$x,y$], $corner$])}
                \State  $dummy\_point \gets$ $\Bigl[\frac{x+corner[0]}{2}$ , $\frac{y+corner[1]}{2}\Bigr]  $ 
                \State $dumy\_rec \gets [0, rec[0], dummy\_point[0],dummy\_point[1]]$
                \State append ($dumy\_rec)$ to $file\_dummy$  
            \EndIf
    \EndWhile    
\EndWhile       

\State $tr$.close()
\State $tr\_dummy$.close()
 
\end{algorithmic}
\end{algorithm*}

Based on the definition of $\mathcal{L}_i$ presented earlier, the determination of each \(i\)'s regional boundary is contingent upon \(\mathcal{N}_i\) at \(f_t\) ($\mathcal{N}_i(f_t)$). 
Nonetheless, this definition might not always guarantee the inclusion of every \( i \) within their respective local region. This can be particularly evident when \( i \) at \( f_t \) lacks neighboring points from all directions, exemplified by person \( 37 \) in~\cref{fig:dummyxamples}a.

To address this issue, we introduce a step to generate dummy points. This involves adding points around each \(i\) at \(f_t\) in areas where they lack direct neighbors. This ensures every \(i\) remains encompassed within their local regions, as illustrated by person \(37\) in~\cref{fig:dummyxamples}c. For this purpose, as depicted in~\cref{fig:dummyxamples}b and~\cref{alg:dummypoints}, firstly, this step involves reading the trajectory data of $i$ that corresponds to $f_t$ (\cref{alg:dummypoints}, lines 1-8). Concurrently, the area surrounding every  $i$  is divided into four equal square regions, each can accommodate at least one $i$ (\cref{alg:dummypoints}, lines 9-17). The location $\langle x,y\rangle_i$ corresponds to the first 2D coordinate of each region (\cref{alg:dummypoints}, lines 12-13). In contrast, the remaining 2D coordinates ($\langle x1,y1\rangle$,$\langle x2,y2\rangle$,$\langle x3,y3\rangle$, $\langle x4,y4\rangle$) required for identifying the regions can be determined by: 
\begin{equation}
\begin{aligned}
\langle x1,y1 \rangle=\langle x-r,y+r \rangle \\
\langle x2,y2 \rangle=\langle x+r,y+r \rangle
\\
\langle x3,y3 \rangle=\langle x+r,y-r \rangle
\\
\langle x4,y4 \rangle=\langle x-r,y-r \rangle,
\end{aligned}
\label{eq:regionscorner}
\end{equation}
where $r$ is the dimension of each square region. Subsequently, each region is checked to verify if it has any pedestrians. In case a region is empty, a dummy point in its center is appended to the input trajectory data. \cref{fig:dummyxamples}b illustrates an example of four regions surrounding person $37$ and two dummy points (yellow dots in first and second empty regions), see  \cref{alg:dummypoints}, lines 18-24.  
After generating the dummy points for all $i$ at $f_t$, the trajectory data is forwarded to the next step, direct neighbor identification.  ~\cref{fig:dummyxamples}c shows a crowd with dummy points in a single $f_t$.

The third step, direct neighbor identification, employs a combination of Voronoi Diagram~\cite{green1978computing} and Convex Hull~\cite{andrew1979another} to find $\mathcal{N}_i(f_t)$ from the input trajectory data with dummy points. 
A Voronoi Diagram is a method for partitioning a plane into several polygonal regions (named Voronoi cells $\mathcal{V}$s)  based on a set of objects/points (called sites)~\cite{green1978computing}. Each $\mathcal{V}$ contains edges and vertices, which form its boundary. \cref{fig:voronoi}a depicts an example of a Voronoi Diagram for $51 \, \mathcal{V}s$ of $51$ sites, where black and yellow dots denote the sites. In the same figure, the set of sites contains $\langle x, y \rangle_i$ (dummy points are included) at a specific $f_t$,  then each $\mathcal{V}_i$ includes only one site $\langle x, y \rangle_i$, and all points within $\mathcal{V}_i$ are closer to site $\langle x, y \rangle_i$ than any other sites $\langle x, y \rangle_q$. Where $q \in $  all $i$ at that $f_t$, and $q \neq i$.   
 
Furthermore, $\mathcal{V}_i$ and $\mathcal{V}_q$ at $f_t$ are considered adjacent if they share at least one edge or two vertices.
For instance, as seen in~\cref{fig:voronoi}, $\mathcal{V}_4$ and $\mathcal{V}_{34}$ are adjacent, while $\mathcal{V}_4$ and $\mathcal{V}_3$ are not adjacent. Since the Voronoi Diagram contains unbounded cells, determining the adjacent cells for each $\mathcal{V}_i$ at $f_t$ may yield inaccurate results. For instance,  most cells of yellow points, which are located at the scene's borders,  are unbounded cells,  as depicted in \cref{fig:voronoi}a. For further clarity, $\mathcal{V}_i(f_t)$  becomes unbounded when $i$ is a vertex of the convex hull that includes all instances of $i$ at $f_t$. As a result, the Voronoi Diagram may not provide accurate results when determining adjacent cells, which is a crucial factor in identifying $\mathcal{N}_i(f_t)$. To overcome such limitation,  Convex Hull~\cite{andrew1979another} is utilized to finite the Voronoi Diagram (unbounded cells) as shown in~\cref{fig:voronoi}b. The Convex Hull is the minimum convex shape that encompasses a given set of points, forming a polygon that connects the outermost points of the set while ensuring that all internal angles are less than $180^{\circ}$~\cite{baillo2021statistical}. For this purpose,  the intersection of each $\mathcal{V}_i(f_t)$ with Convex Hull of all $i$ at  $f_t$ is calculated, then the $\mathcal{V}_i(f_t)$ in the diagram are updated based on the intersections to obtain the bounded Voronoi Diagram of all $i$ at $f_t$ (\cref{alg:dnidentificationstep}, lines 5-12). In more details, the Convex Hull of all $i$ at $f_t$ is measured (\cref{alg:dnidentificationstep}, line 8). After that, the intersection between $\mathcal{V}_i(f_t)$ and the Convex Hull at $f_t$  is computed. And finally, we update the Voronoi Diagram at each $f_t$ using the calculated interactions to obtain the corresponding bounded one as shown in~\cref{fig:voronoi}b (\cref{alg:dnidentificationstep}, lines 8-11). After creating the bounded Voronoi Diagram, individuals in the direct adjacent Voronoi cells of  $\mathcal{V}_i(f_t)$ are $\mathcal{N}_i(f_t)$, (\cref{alg:dnidentificationstep}, lines 12-20). For example, in~\cref{fig:voronoi}b, direct adjacent Voronoi cells of $\mathcal{V}_3$ at $f_t$ are $\{\mathcal{V}_2, \mathcal{V}_{22}, \mathcal{V}_{35}, \mathcal{V}_{15}, \mathcal{V}_7  \}$, and  $\mathcal{N}_3 =\{2, 22, 35, 15, 7 \}$.

The last step, local region extraction, aims to extract the local region of each $i$ at $f_t$, where $i \notin$ dummy points.  
The step firstly finds  $\mathcal{L}_i(f_t)$  based on each $\langle x,y \rangle_j$, where $j\in \mathcal{N}_i(f_t)$, 
\cref{fig:dummyxamples}c.   Then,    $\mathcal{L}_i(f_t)$ are cropped from corresponding $f_t$ and passed to the next module, which will be discussed in the next section. \cref{fig:LR-crowdexamples}b displays examples of cropped local regions.
 
\begin{algorithm*}[h]
\caption{Pseudo code of direct neighbor identification step}\label{alg:dnidentificationstep}
 \textbf{Inputs:} \\
 \hspace*{\algorithmicindent} \textbf{$tr\_dummy$}: a file of pedestrian trajectory data (over seconds) with dummy points.\\ 
 \textbf{Outputs:} \\
 \hspace*{\algorithmicindent} \textbf{$direct\_neighbor$}: a file of direct neighbors for pedestrians over seconds.  
\begin{algorithmic}[1]

\State $file \gets $ open($tr\_dummy$)
\State $file\_dn \gets $ open($direct\_neighbor$)
\State $data \gets$ load($file$)
\State $frames \gets$ unique($data[:,0]$)
 
\While{$fr$ in  $frames$}
    \State $data\_fr \gets$ filter($data$ , $fr$)
    \State $vor\_diagram \gets$ Voronoi($data\_fr[:,2:4]$)

    \State $CH \gets$ ConvexHull($data\_fr[:,2:4]$)
    \ForEach{$region \in vor\_diagram$.regions}
        \State $vor\_diagram .region \gets  vor\_diagram.region \, \cap  \, CH$
    \EndFor
  
    \State $cells \gets vor\_diagram$.regions
    \While{$i$ in $data\_fr[:,0]$ } 
        \State $cell\gets cells[i]$ 
        \State $dn\_cells \gets $ find\_direct\_neighbor\_cells ($cell$)
        \State $dn\_i \gets$ $dn\_cells$.sites
        \While{$a\_j$ in $dn\_i$}
        \State $rec \gets [fr, i, j]$
        \State write ($rec)$ to $file\_dn$  
        \EndWhile
    \EndWhile
\EndWhile
\State $file$.close()
\State $file\_dn$.close()
\end{algorithmic}
\end{algorithm*}

\subsubsection{EfficientNetV1B0-based Deep Feature Extraction}

 To extract the deep features from each individual's behavior, the feature extraction part of EfficientNetV1B0 is trained on their local regions $\mathcal{L}_i(f_t)$.  EfficientNetV1B0 is a CNN architecture that has gained popularity for various computer vision tasks due to its efficient use of resources and fewer parameters than other state-of-the-art models~\cite{tan2019efficientnet}. Furthermore, it has achieved high accuracy on multiple image classification datasets. Additionally, The experiments in this article~(\cref{comparisons-baselines-cnn}) indicate that combining EfficientNetV1B0 with local regions yields the highest accuracy compared to other popular CNN architectures integrated with local regions. Therefore, EfficientNetV1B0's feature extraction part is employed to find more helpful information from each individual's behavior.

\begin{figure}[h]
   \centering
   \includegraphics[width=0.5\textwidth]{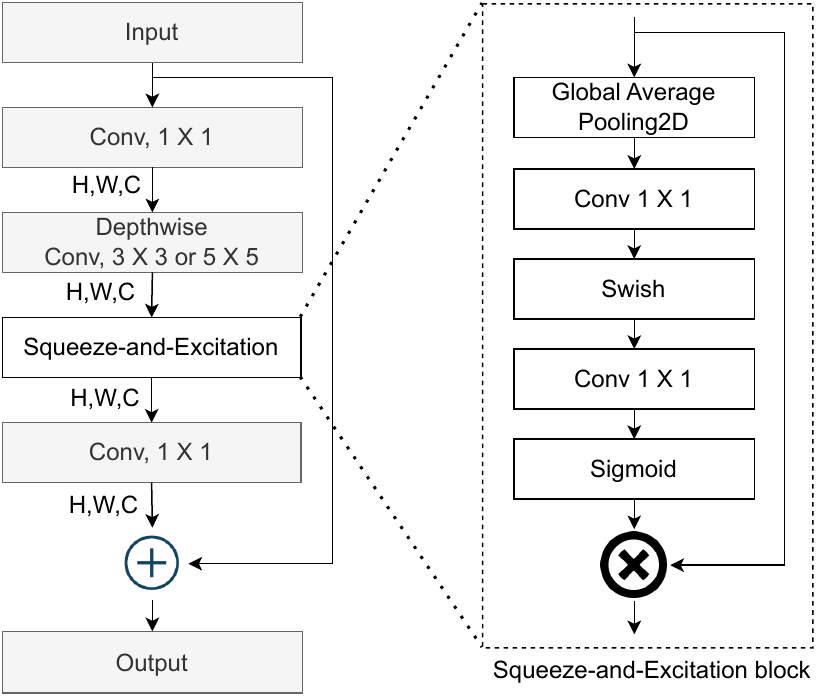}
  \caption{ The architecture of MBConv block.}
  \label{fig:MBConv-SE}
\end{figure}

  The architecture of the efficientNetV1B0-based deep feature extraction model is depicted in \cref{fig:framwork}. Firstly, it applies a $3 \times 3$ convolution operation to the input image, a local region with dimensions of $224 \times 224 \times 3$. Following this, 16 mobile inverted bottleneck convolution (MBConv) blocks~\cite{sandler2018mobilenetv2} are employed to extract deep features (feature maps) $\in \mathbb{R}^{7 \times 7 \times 320}$ from each $\mathcal{L}_i(f_t)$. In more detail, the MBConv blocks used consist of one MBConv1, $3 \times 3$,  six MBConv6, $3 \times 3$, and nine MBConv6, $5 \times 5$. \cref{fig:MBConv-SE} illustrates the structure of the MBConv block, which employs a $1 \times 1$ convolution operation to expand the depth of feature maps and capture more information. A $3 \times 3$ depthwise convolution follows this to decrease the computational complexity and number of parameters. Additionally, batch normalization and swish activation~\cite{ramachandran2017searching} are applied after each convolution operation. The MBConv block then employs a Squeeze-and-Excitation block~\cite{hu2018squeeze} to enhance the architecture's representation power. The Squeeze-and-Excitation block initially performs global average pooling to reduce the channel dimension. Then it applies an excitation operation with Swish~\cite{ramachandran2017searching} and Sigmoid~\cite{han1995influence} activations to learn channel-wise attention weights. These weights represent the significance of each feature map and are multiplied by the original feature maps to generate the output feature maps. After the Squeeze-and-Excitation block, another $1 \times 1$ convolution with batch normalization is used to reduce the output feature maps' dimensionality, resulting in the final output of the MBConv block.

  The main difference between MBConv6 and MBConv1 is the depth of the block and the number of operations performed in each block; MBConv6 is six times that of MBConv1. Note that MBConv6, $5 \times 5$ performs the identical operations as MBConv6, $3  \times 3$, but MBConv6, $5  \times 5$ applies a kernel size of $5  \times 5$, while MBConv6, $3  \times 3$ uses a kernel size of  $3  \times 3$.

\subsection{Labeling Component}
The objective of the labeling component is to analyze the feature maps obtained from the previous component and identify the pushing individuals in the input video. This is accomplished through a binary classification task, followed by an annotation process. To carry out the classification task, as shown in~\cref{fig:framwork}, a $1 \times 1$ convolution operation, global average pooling2D, a fully connected layer, and a Sigmoid activation function are combined. 
A $1 \times 1$ convolutional operation is used to increase the number of channels in feature maps, leading to more information. The new dimension of feature maps for each $\mathcal{L}_i(f_t)$ is $7 \times 7 \times 1280 $. After that,  the global average pooling2D is utilized to transform the feature maps to $1 \times 1 \times 1280$ and feed them to the fully connected layer. Then,  the fully
connected layer with a Sigmoid activation function finds
the probability $\delta $ of the pushing label for the corresponding $i$ at $f_t$. 
Finally, the classifier uses   threshold  to identify the class of $i$ at $f_t$ as~\cref{eq:class}:
\begin{equation}
\label{eq:class}
Class ({i,f_t}) =
\begin{cases}
\text{pushing} & \text{if } \delta \geq threshold
\\
\text{non-pushing} & \text{if } \delta < threshold
 
\end{cases}
\end{equation}

\begin{figure*}[t]
   \centering
   \includegraphics[width=1\textwidth]{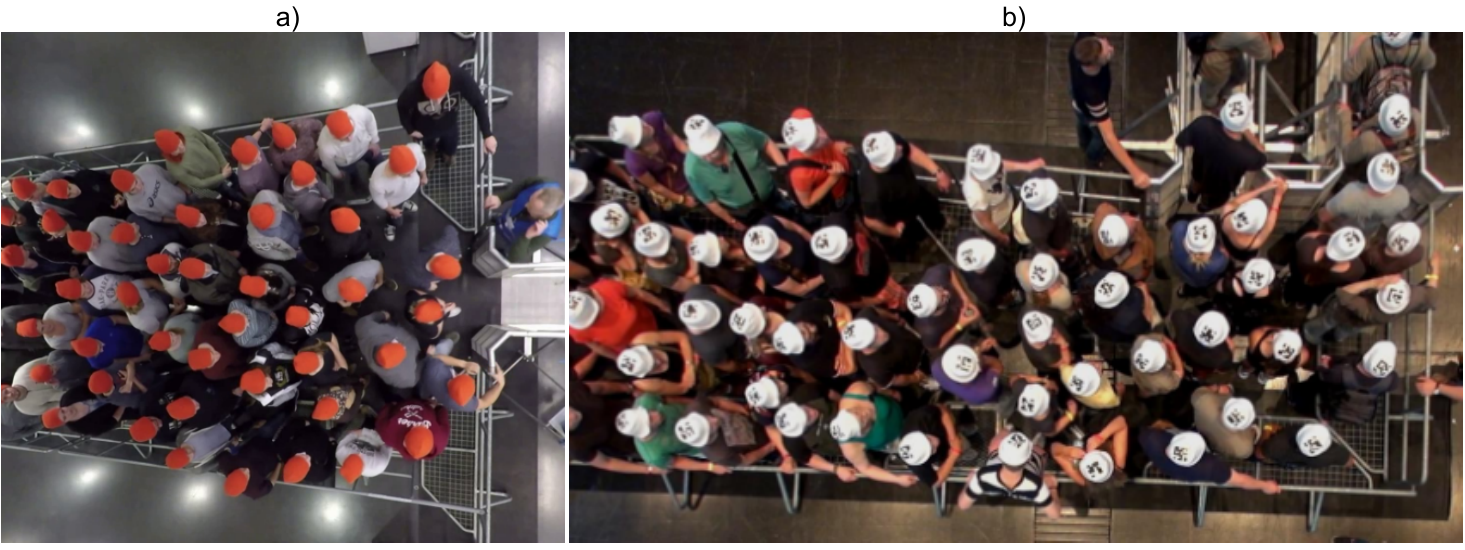}
  \caption{Overhead view of exemplary experiments. a) Experiment 270, as well as Experiments 50, 110, 150, and 280 used the same setup but with different widths of the entrance area ranging from 1.2 to 5.6 m based on the experiment~\cite{crowdqueue}.  b) Experiment entrance\_2~\cite{entrance2}  The entrance gate's width is 0.5 m in all setups.}
  \label{fig:overhead-view}
\end{figure*}

\begin{table*}[t]
 \caption{Characteristics of the chosen experiments.}
  \centering
  \begin{tabular} {c c c c c c}
\toprule
Video experiment * & 
Width (m) & 
Pedestrian total &
Number of gates &
Duration (s) & 
Resolution
\\
\midrule
50 & 
4.5 &
42 &
1 &
37&
1920 $\times$  1440\\

110 & 
1.2 &
63 &
1 &
53& 
1920 $\times$  1440\\

150 & 
5.6 &
57 &
1 &
57 &
1920 $\times$  1440\\

270 & 
3.4 &
67 &
1 &
59&
1920 $\times$  1440\\

280 & 
3.4 &
67 &
1 &
67 &
1920 $\times$  1440\\

Entrance\_2 & 
3.4 &
123 &
2 &
125 &
1920 $\times$  1080\\

\bottomrule
\multicolumn{6}{p{400pt}}{ *The same names as reported in~\cite{crowdqueue,entrance2}. m stands for meter, and s refers to second.}\\
  
\end{tabular}
  \label{tb:exper-char}
\end{table*}

\begin{figure*}[b]
   \centering
   \includegraphics[width=1\textwidth]{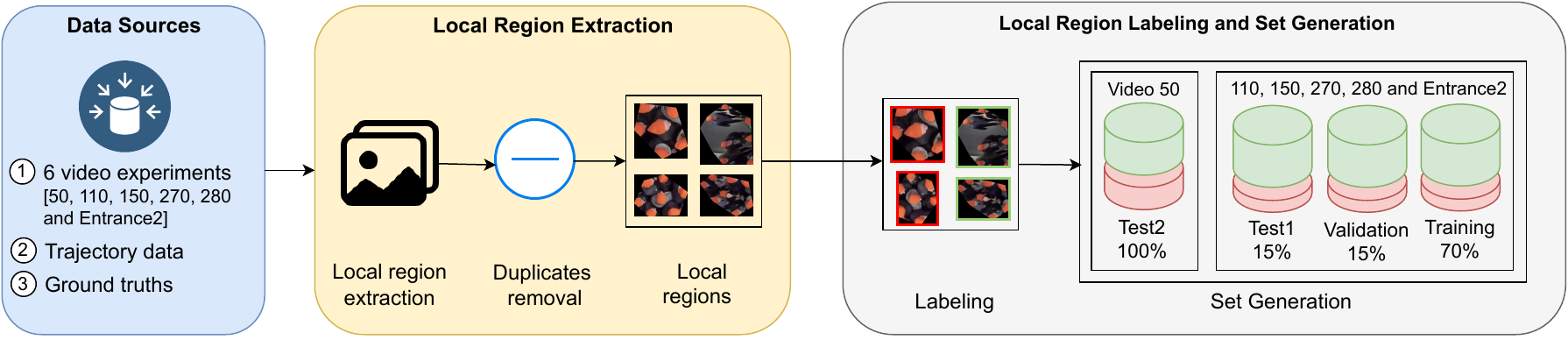}
  \caption{Pipeline of dataset preparation. In the part 'Local Region Labeling and Set Generation', red refers to the pushing class and pushing sample, while the non-pushing class and non-pushing sample are represented in green.}
  \label{fig:datasetPreparation}
\end{figure*}

By default, the threshold value for binary classification is set to 0.5, which works well for a dataset with a balanced distribution. Unfortunately, the new pushing dataset created in~\cref{sec:noveldataset} for training and evaluating the proposed framework is imbalanced, and using the default threshold may lead to poor performance of the introduced trained classifier on that dataset~\cite{esposito2021ghost}. Therefore, adjusting the threshold in the trained classifier is required to obtain better accuracy for both pushing and non-pushing classes. The methodology for finding the optimal threshold for the classifier will be explained in detail in~\cref{sec:metricandoptimalthreshold}.  Following training and adjusting the classifier's threshold, it can categorize individuals $i$ as pushing or non-pushing. At the same time, the annotation process draws a red circle around the head of each pushing person in the corresponding frames $f_t$ and finally generates an annotated video. 

The following section will discuss the training and evaluating processes of the propsed framework. 

\section{Training and Evaluating the Framework}
\label{trainingandevaluatingtheframework}
This section introduces a novel labeled dataset, as well as presents the parameter setups for the training process, evaluation metrics, and the methodology for improving the framework's performance on an imbalanced dataset.

\subsection{A Novel Dataset Preparation}
\label{sec:noveldataset}

Here, it is aimed to create the labeled dataset for training and evaluating the proposed framework. The dataset consists of a training set, a validation set for the learning process, and two test sets for the evaluation process.  These sets comprise $\mathcal{L}_i(f_t)$ labeled as either pushing or non-pushing. In this context, each pushing $\mathcal{L}_i(f_t)$ means $i$ at $f_t$ contributes pushing, while every non-pushing $\mathcal{L}_i(f_t)$ indicates that $i$ at $f_t$ follows the social norm of queuing.  The following will discuss the data sources and methodology used to prepare the sets.

The dataset preparation is based on three data sources: 1) Six videos of real-world experiments of crowded event entrances. 2) Pedestrian trajectory data. 3) Ground truths for pushing behavior. Six video recordings of experiments with their corresponding pedestrian trajectory data are selected from the data archive hosted by Forschungszentrum J\"ulich~\cite{crowdqueue, entrance2}. This data is licensed under CC Attribution 4.0 International license.  The experimental situations mimic the crowded event entrances, and static top-view cameras were employed to record the experiments with a frame rate of 25 frames per second. For more clarity,  \cref{fig:overhead-view} shows overhead views of exemplary experiments, and \cref{tb:exper-char} summarizes the various characteristics of the chosen experiments. Additionally, ground truth labels constructed by the manual rating system~\cite{usten2022pushing} are used for the last data source. In this system, social psychologists observe and analyze video experiments frame-by-frame to manually identify individuals who are pushing over time.  The experts use PeTrack software~\cite{boltes2010automatic} to manage the manual tracking process and generate the annotations as a text file. For further details on the manual system, readers can refer to Ref.~\cite{usten2022pushing}. 

\begin{table*}[h]
\caption{Summary of the prepared sets.}

\footnotesize
\begin{tabular}{lccccccccccccc}
\toprule
 &
  \multicolumn{3}{c}{Number of samples} &
  \multicolumn{2}{l}{Labeled dataset} &
  \multicolumn{2}{l}{Training set} &
  \multicolumn{2}{l}{Validation set} &
  \multicolumn{2}{l}{Test set1} &
  \multicolumn{2}{l}{Test set2} \\ \midrule
 Video  & Original & Deleted & Distinct & P    & NP   & P    & NP   & P   & NP   & P   & NP   & P   & NP  \\ \midrule
110                     & 1046     & 1       & 1045          & 548  & 497  & 365  & 331  & 99  & 84   & 84  & 82   &     &     \\
150                     & 1469     & 70      & 1399          & 625  & 774  & 455  & 547  & 83  & 113  & 87  & 114  &     &     \\
270                     & 1627     & 11      & 1616          & 577  & 1039 & 401  & 727  & 84  & 161  & 92  & 151  &     &     \\
280                     & 1822     & 44      & 1778          & 287  & 1491 & 213  & 1104 & 44  & 181  & 30  & 206  &     &     \\
Entrance\_2             & 6204     & 325     & 5879          & 1030 & 4849 & 726  & 3403 & 156 & 715  & 148 & 731  &     &     \\
\rowcolor[HTML]{eeeeee} 
Total                   & 12168    & 451     & 11717         & 3067 & 8650 & 2160 & 6112 & 466 & 1254 & 441 & 1284 &     &     \\
\rowcolor[HTML]{eeeeee} 
50*                      &          &         &               & 317  & 344  &      &      &     &      &     &      & 317 & 344 \\ \bottomrule
\multicolumn{14}{p{450pt}}{* Video 50 is used exclusively for the evaluation process, while the remaining video experiments will be employed for both training and evaluation.}\\
\end{tabular}
\label{tab:lableddataset}
\end{table*}

Here, the methodology used for papering the dataset is described. As shown in~\cref{fig:datasetPreparation},  it consists of two phases: local region extraction; and local region labeling and set generation. The first phase aims to extract local regions (samples) from videos while avoiding duplicates.  To accomplish this, the phase initially extracts frames from the input videos second by second. It employs After that the Voronoi-based local region extraction module to identify and crop the samples from the extracted frames. \cref{tab:lableddataset} demonstrates the number of extracted samples from each video, and \cref{fig:LR-crowdexamples}b shows several examples of local regions. Preventing the presence of duplicate samples between the training, validation, and test sets is crucial to obtain a reliable evaluation for the model. Therefore, this phase removes similar and slightly different samples before proceeding to the next phase. It involves using a pre-trained MobileNet CNN model to extract deep features/embeddings from the samples and cosine similarity to find duplicate or near duplicate samples based on their features~\cite{deduplicate}. This technique is more robust than comparing pixel values, which can be sensitive to noise and lighting variations~\cite{zheng2016improving}. \cref{tab:lableddataset} depicts the number of removed duplicate samples.

On the other hand, the local region and set generation phase is responsible for labeling the extracted samples and producing the sets, including one training set, one validation set, and two test sets. This phase utilizes the ground truth label of each $i$ at $f_t$ to label the samples ($\mathcal{L}_i(f_t)$). If $i$ at $f_t$ contributing to pushing, $\mathcal{L}_i(f_t)$ is categorized as pushing; otherwise,
it is classified as non-pushing.  Examples of pushing samples can be found in~\cref{fig:LR-crowdexamples}b. The generated labeled dataset from all video experiments comprises 3384 pushing samples and 8994 non-pushing samples. The number of extracted pushing and non-pushing samples from each video is illustrated in~\cref{tab:lableddataset}. After creating the labeled dataset, the sets are generated from the dataset. Specifically, the second phase randomly divides the extracted frames from video experiments 110, 150, 270, 280, and Entrance\_2 into three sets: \qty{70}{\percent}, \qty{15}{\percent}, and \qty{15}{\percent} for training, validation, and test sets, respectively.  Then, using these sets, it generates the training, validation, and test (test set 1) sets from the labeled corresponding samples ($\mathcal{L}_i(f_t)$). Another test set (test set 2) is also developed from the labeled samples extracted from the complete video experiment 50. \cref{tab:lableddataset} shows the summary of the generated sets.


To summarize,  four labeled sets were created: the training set, which consists of 2160 pushing samples and 6112 non-pushing samples; the validation set, which contains 466 pushing samples and 1254 non-pushing samples; the test set 1, which includes 441 pushing samples and 1284 non-pushing samples; and the test set 2, comprising 317 pushing samples and 344 non-pushing samples.

\subsection{Parameter Setup}

\cref{tab:hyperparameter} shows parameters used during the training process. They were chosen based on experimentation to obtain optimal performance with the new
dataset. To prevent overfitting,  the training was halted if the validation accuracy did not improve after 20 epochs.

\begin{table}[h]
\centering

\caption{The hyperparameter values used in the training process.}
\label{tab:hyperparameter}
\setlength{\tabcolsep}{3pt}

\begin{tabular}{p{100pt}p{100pt}}  \hline
Parameter     & Value                \\ \hline
Optimizer     & Adam                 \\
Loss function & Binary cross-entropy \\
Learning rate & 0.001                \\
Batch size    & 32                   \\
Epoch         & 100\\  \hline              
\end{tabular}
\end{table}

\subsection{Evaluation Metrics and Performance Improvement}
\label{sec:metricandoptimalthreshold}

This section will discuss the metrics chosen for evaluating the performance of the proposed framework. Additionally, it will explore the methodology employed to enhance the performance of the trained imbalanced classifier, thereby improving the overall effectiveness of the framework.

Given the imbalanced distribution of the generated local region dataset, the framework exhibits a bias towards the majority class (non-pushing). Consequently, it becomes crucial to employ appropriate metrics for evaluating the performance of the imbalanced classifier. As a result, a combination of metrics was adopted, including macro accuracy, True Pushing Rate (TPR), True Non-Pushing Rate (TNPR), and Area Under the receiver operating characteristic Curve (AUC) on both test set 1 and test set 2. The following provides a detailed explanation of these metrics.

TPR, also known as sensitivity, is the ratio of correctly classified pushing samples to all pushing samples, and it is defined as:

\begin{equation}
\label{eq:tpr}
TPR=\frac{TP}{TP+FNP},   
\end{equation}
where TP and FNP denote correctly classified pushing persons and incorrectly predicted non-pushing persons.

TNPR, also known as specificity, is the ratio of correctly classified non-pushing samples to all non-pushing samples, and it is described as:

\begin{equation}
\label{eq:tnpr}
TNPR=\frac{TNP}{TNP+FP},   
\end{equation}
where TNP and FP stand for correctly classified non-pushing persons and incorrectly predicted pushing persons.

\begin{figure*}[t!]
   \centering
   \includegraphics[width=0.9\textwidth]{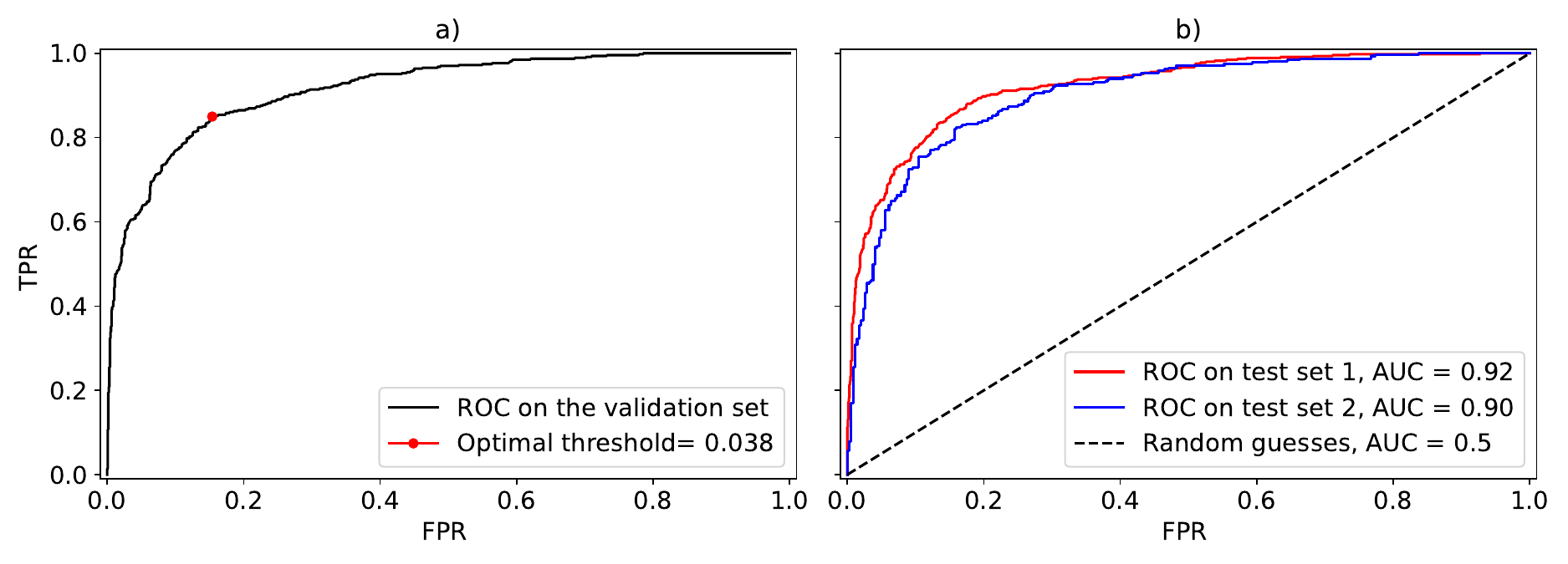}
  \caption{ROC curves for the introduced framework. a) ROC curve with an optimal threshold on the validation set. b) ROC curves with AUC values on test set 1 and test set 2. TPR stands for true pushing rate, while FPR refers to false pushing rate. }
  \label{fig:effb0thresholdauc}
\end{figure*}

Macro accuracy, or balanced accuracy,  is the average proportion of correct predictions for each class individually. This metric ensures that each class is given equal significance, irrespective of its size or distribution within the dataset. For more clarity, it is just the average of TPR and TNPR as:

\begin{equation}
\label{eq:macroaccuracy}
Macro \, accuracy=\frac{TPR + TNPR}{2}.   
\end{equation}

\begin{figure*}[t]
   \centering
   \includegraphics[width=0.65\textwidth]{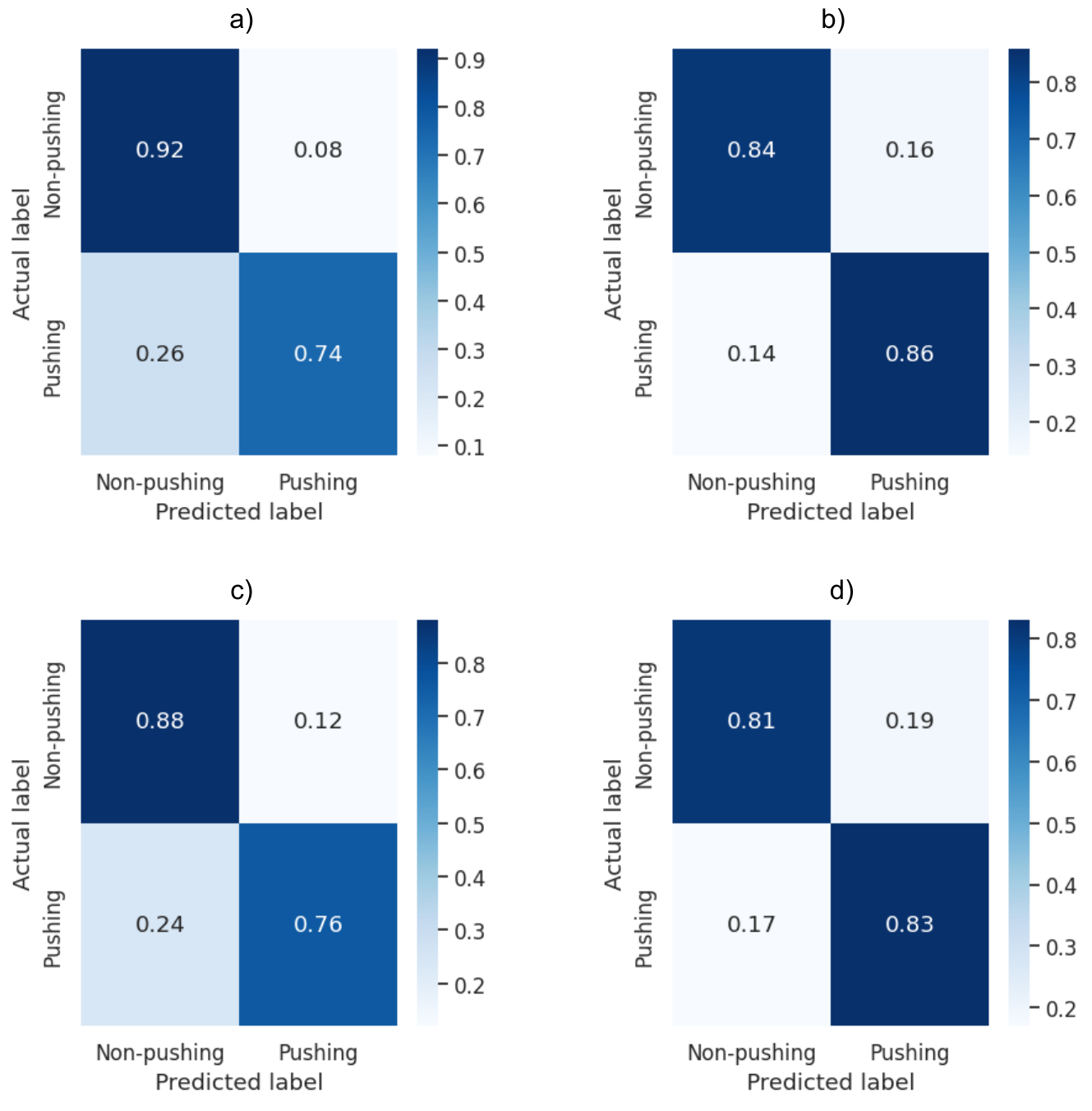}
  \caption{Confusion matrix of the proposed framework on a) Test set 1 with default threshold. b) Test set 1 with the optimal threshold. c) Test set 2  with default threshold. d) Test set 2 with the optimal threshold.  }
  \label{fig:effb0-cm}
\end{figure*}

\begin{table*}[h!]
\centering
\footnotesize
\caption{Performance of the proposed framework on both test sets.}
\begin{tabular}{lcccccccc} 
\toprule
                                   & \multicolumn{4}{c}{Test set 1 \%} & \multicolumn{4}{c}{Test set 2 \%} \\   \cmidrule(r){2-5}   \cmidrule(r){6-9}
Threshold &
  \multicolumn{1}{l}{Macro accuracy } &
  \multicolumn{1}{l}{TNPR  } &
  \multicolumn{1}{l}{TPR } &
  \multicolumn{1}{l}{(TPR-TNPR)} &
  \multicolumn{1}{l}{Macro accuracy  } &
  \multicolumn{1}{l}{TNPR  } &
  \multicolumn{1}{l}{TPR  } &
  \multicolumn{1}{l}{(TPR-TNPR)} \\ \midrule
Default: 0.5                       & 83    & 92    & 74    & 18    & 82    & 88    & 76    & 12    \\
\multicolumn{1}{c}{\textbf{Optimal: 0.038}} & \textbf{85}    & 84    & 86   & \textbf{2}     & 82    & 81    & 83    & 2 \\ \bottomrule 
\multicolumn{9}{p{450pt}}{TNPR and TPR are true non-pushing rate and true pushing rate, respectively.}\\  
\end{tabular}
\label{tab:ourframeworkperfromance}
\end{table*}

AUC is a metric that represents the area under the Receiver Operating Characteristics (ROC) curve. The ROC curve illustrates the performance of a classification model across various threshold values. It plots the false positive rate (FPR) on the horizontal axis against the true positive rate (TPR) on the vertical axis. AUC values range from 0 to 1, where a perfect model achieves an AUC of 1, while a value of 0.5 indicates that the model performs no better than random guessing~\cite{devries2021using}. \cref{fig:effb0thresholdauc}a shows an example of a ROC curve with AUC value. 

As mentioned above, the binary classifier employs a threshold to convert the calculated probability into a predicted class. The pushing class is predicted if the probability exceeds the threshold; otherwise, the non-pushing label is predicted. The default threshold is typically set at 0.5. However, this value leads to poor performance of the introduced framework because EfficientNetV1B0 and classification were trained on imbalanced dataset~\cite{esposito2021ghost}. In other words, the default threshold yields a high TNPR and a low TPR in the framework. To address the imbalance issue and enhance the framework's performance, it becomes necessary to determine an optimal threshold that achieves a better balance between TPR and   FPR (1-TNPR). To accomplish this,  the ROC curve is utilized over the validation set to identify the threshold value that maximizes TPR and minimizes FPR.    Firstly, TPR and TNPR are calculated for several thresholds ranging from 0 to 1. Then, the threshold that yields the minimum value for the following objective function (\cref{eq:objectivefunction} is considered the optimal threshold:  

\begin{equation}
\label{eq:objectivefunction}
Objective \, function=\lvert TPR-TNPR \rvert.   
\end{equation}

As shown in~\cref{fig:effb0thresholdauc}a, the red point refers to the optimal threshold of the classifier used in the proposed framework, which is 0.038.

\section{Evaluation and Results }
\label{sec:evaluationandresults}

Here, several experiments were conducted to evaluate the performance of the proposed framework. Initially, the performance of the proposed framework itself is assessed. Subsequently, It is compared with five other CNN-based frameworks. The influence of the deep feature extraction module on the proposed framework's performance is also investigated. Finally,  the impact of the local region extraction module on the framework's performance is explored. All experiments and implementations were performed on Google Colaboratory Pro, utilizing Python 3 programming language with Keras, TensorFlow 2.0, and OpenCV libraries. In Google Colaboratory Pro, the hardware setup comprises an NVIDIA GPU with a 15 GB capacity and a system RAM of 12.7 GB. Moreover, the framework and all the baselines developed for comparison in the experiments were trained using the same sets~(\cref{tab:lableddataset}) and hyperparameter values (\cref{tab:hyperparameter}).  

\subsection{Performance of the Proposed Framework}

The performance of the proposed framework was evaluated using the generated dataset (\cref{tab:lableddataset}) and various metrics, including macro accuracy, TPR, TNPR, and AUC. We first trained the proposed framework's EfficientNetB0-based deep feature extraction module and labeling component on the training and validation sets. Subsequently,  the framework's performance on test set 1 and test set 2 were assessed.

\cref{tab:ourframeworkperfromance} shows that the introduced framework, with the default threshold, obtained macro accuracy of \qty{83}{\percent},  TPR of \qty{74}{\percent}, and  TNPR of \qty{92}{\percent} on test set 1. On the other hand, it achieved \qty{82}{\percent} macro accuracy, \qty{88}{\percent} TNPR, and \qty{76}{\percent} TPR on test set 2. However, it is clear that the TPR is significantly lower than the TNPR on both test sets, see~\cref{fig:effb0-cm}a and c. To balance the TPR and TNPR and improve the TPR, the optimal threshold is 0.038, as shown in~\cref{fig:effb0thresholdauc}a. This threshold increases TPR by \qty{12}{\percent} and \qty{7}{\percent} on test set 1 and test set 2, respectively, without affecting the accuracy, see~\cref{fig:effb0-cm}b and d. In fact, the framework's accuracy improved by \qty{2}{\percent} on test set 1. The ROC curves with AUC values for the framework on the two test sets are shown in~\cref{fig:effb0thresholdauc}b, with AUC values of 0.92 and 0.9 on test set 1 and test set 2, respectively.

To summarize, with the optimal threshold, the proposed framework achieved an accuracy of \qty{85}{\percent}, TPR of \qty{86}{\percent}, and TNPR of \qty{84}{\percent} on test set 1, while obtaining \qty{82}{\percent} accuracy, \qty{81}{\percent} TPR, and \qty{83}{\percent} TNPR on test set 2. The next section will compare the framework's performance with five baseline systems for further evaluation.

\subsection{Comparison with Baseline CNN-based Frameworks}
\label{comparisons-baselines-cnn}

In this section, the results of further empirical comparisons are shown to evaluate the framework's performance against five baseline systems. Specifically, it explores the impact of the EfficientNetV1B0-based deep feature extraction module on the overall performance of the framework. To achieve this, EfficientNetV1B0 in the deep feature extraction module of the proposed framework was replaced with other CNN architectures, including EfficientNetV2B0~\cite{tan2021efficientnetv2} (baseline 1), Xception~\cite{chollet2017xception} (baseline 2), DenseNet121~\cite{huang2017densely} (baseline 3), ResNet50~\cite{he2016deep}(baseline 4), and MobileNet~\cite{howard2017mobilenets} (baseline 5). To ensure fair comparisons,  the five baselines were trained and evaluated using the same sets, hyperparameters, and metrics as those used for the proposed framework.

\begin{figure*}[t]
   \centering
   \includegraphics[width=1\textwidth]{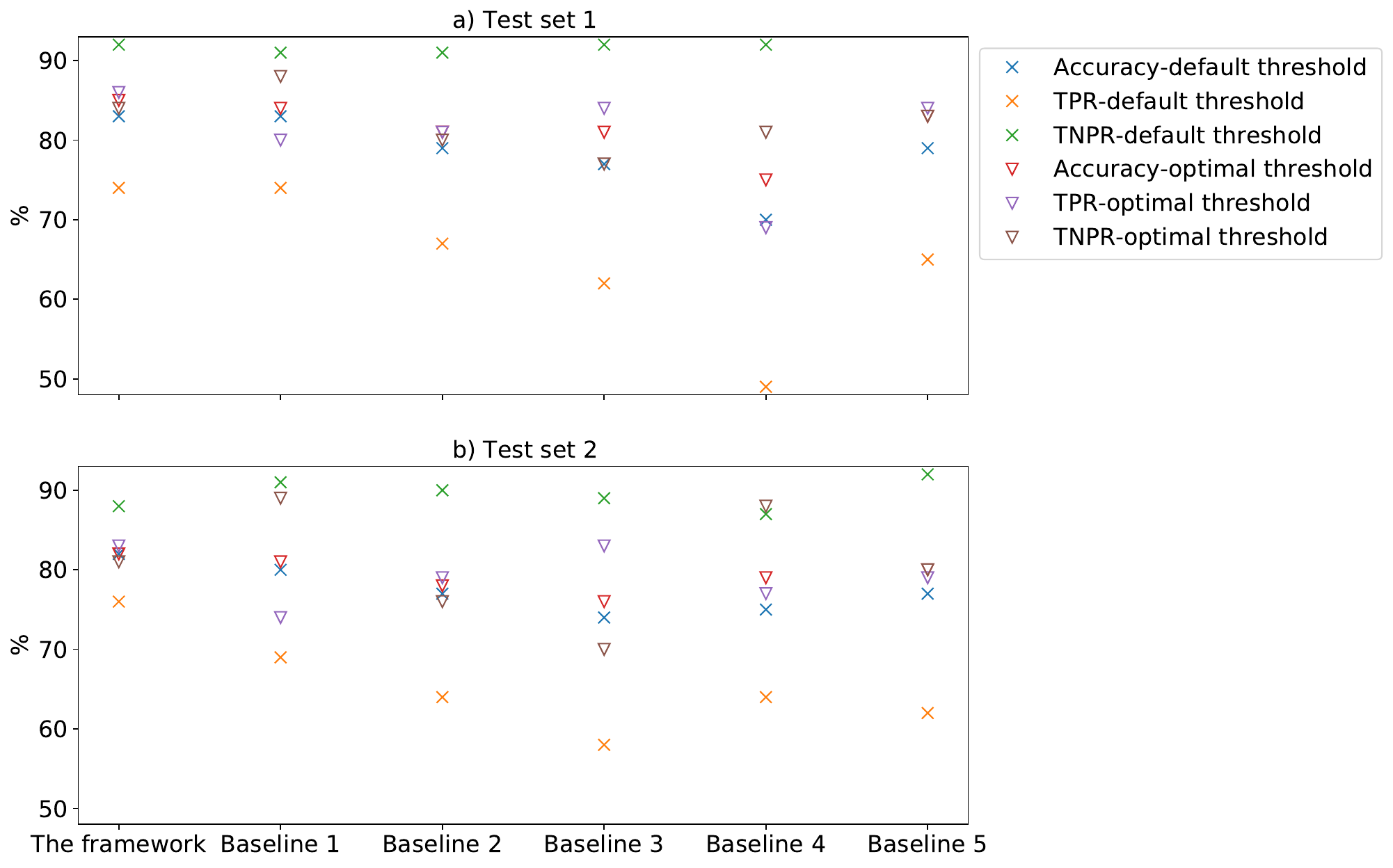}
  \caption{Comparison between the framework (based on EfficientNetV1B0) with the baseline frameworks based on other popular CNN architectures.}
  \label{fig:comparison-optimal-default}
\end{figure*}

\begin{figure*}[b]
   \centering
   \includegraphics[width=0.9\textwidth]{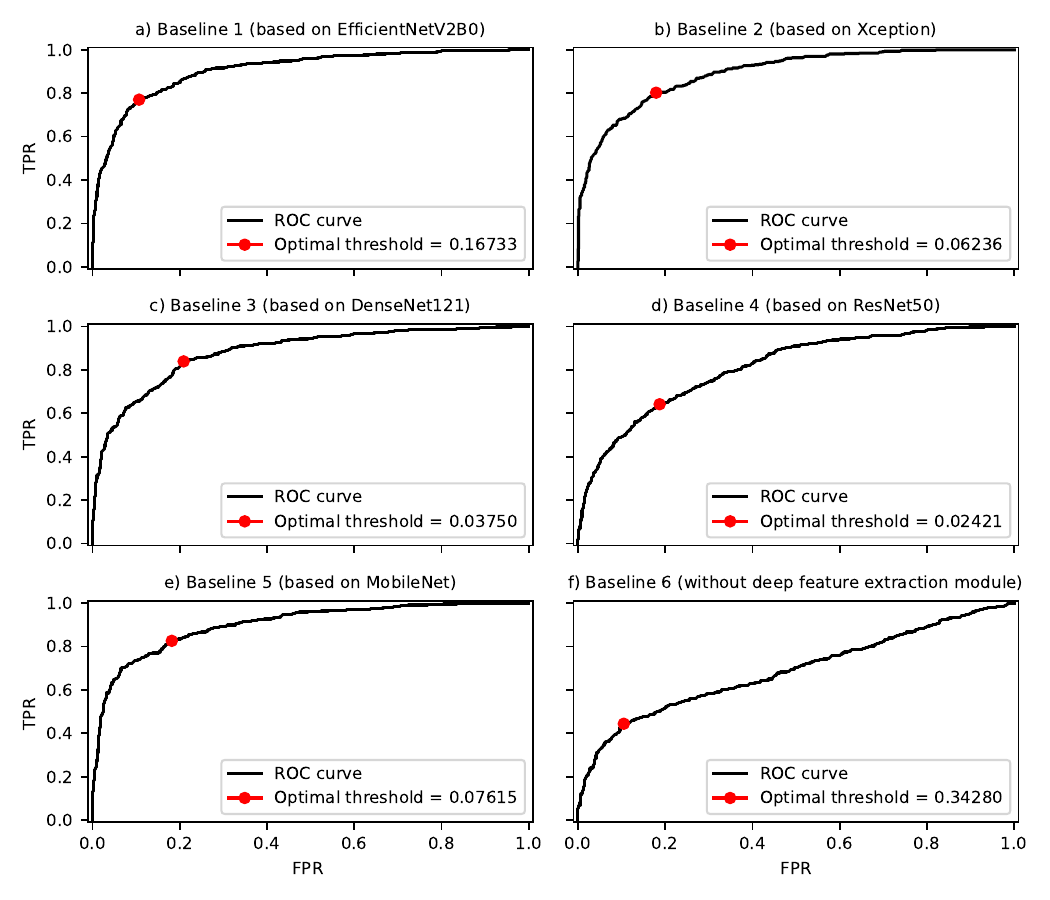}
  \caption{ROC curves with optimal thresholds for the baselines over the validation set. TPR stands for true pushing rate, while FPR refers to false pushing rate.  ROC stands for Receiver Operating Characteristics. }
  \label{fig:thresholds}
\end{figure*}

Before delving into the comparison of the results, it is essential to know that CNN models renowned for their performance on some datasets may perform poorly on others~\cite{yamashita2018convolutional}. This discrepancy becomes more apparent when datasets differ in several aspects, such as size, clarity of relevant features among classes, or overall data quality. Powerful models can be prone to overfitting issues, while simpler models may struggle to capture relevant features in complex datasets with intricate patterns and relationships. Therefore, it's crucial to carefully select or develop an appropriate CNN architecture for a specific issue. For instance, EfficientNetV2B0 demonstrates superior performance compared to EfficientNetV1B0 across various classification tasks~\cite{tan2021efficientnetv2}, including the ImageNet dataset. Moreover, it surpasses the previous version in identifying regions that exhibit pushing persons in motion information maps of crowds~\cite{alia2022hybrid, alia2023cloud}. These remarkable outcomes can be attributed to the efficient blocks employed for feature extraction, namely the Mobile Inverted Residual Bottleneck Convolution~\cite{sandler2018mobilenetv2} and Fused Mobile Inverted Residual Bottleneck Convolution~\cite{gupta2019efficientnet}. Nevertheless, it should be noted that the presence of these efficient blocks does not guarantee the best performance in identifying pushing individuals based on local regions within the framework. Hence, in this section, the impact of six of the most popular and efficient CNN architectures on the performance of the proposed framework was empirically studied. For clarity, EfficientNetV1B0 was used within the framework, while the remaining CNN architectures were employed in the baselines.

\begin{table*}[h]
\centering
\footnotesize
\caption{Comparative analysis of the developed framework and the five CNN-based frameworks.}
\begin{tabular}{lccccccccc}

\toprule 
                 &       &  \multicolumn{4}{c}{Test set 1 \%}     &  \multicolumn{4}{c}{Test set 2 \%}  \\
                   \cmidrule(r){3-6}   \cmidrule(r){7-10}
 Framework   & Threshold & M. acc. & TNPR & TPR & $\lvert$ TPR-TNPR $\rvert$ & M. acc. & TNPR & TPR & $\lvert$ TPR-TNPR $\rvert$  \\ \midrule
The framework   & Default: 0.5   & 83    & 92 & 74 & 18 & 82    & 88 & 76 & 12   \\
                 &Optimal: 0.038 & \textbf{85}    &84 & 86 & \textbf{2}  &\textbf{ 82}    & 81 & 83 & \textbf{2}    \\ \midrule
Baseline 1 & Default: 0.5   & 83    & 91 & 74 & 17 & 80    & 91 & 69 & 22   \\
                 &Optimal: 0.167  & 84    & 88 & 80 & 8  & 81    & 89 & 74 & 15  \\ \midrule
Baseline 2       & Default: 0.5    & 79    & 91 & 67 & 24 & 77    & 90 & 64 & 26   \\
                 & Optimal: 0.062 & 81    & 80 & 81 & 1  & 78    & 76 & 79 & 3    \\ \midrule
Baseline 3      &Default: 0.5    & 77    & 92 & 62 & 30 & 74    & 89 & 58 & 31   \\
                 & Optimal: 0.038 & 81    & 77 & 84 & 7  & 76    & 70 & 83 & 13   \\ \midrule
Baseline 4         & Default: 0.5    & 70    & 92 & 49 & 43 & 75    & 87 & 64 & 23  \\
                 & Optimal: 0.024  & 75    & 81 & 69 & 12 & 79    & 88 & 77 & 11  \\ \midrule
Baseline 5        & Default: 0.5    & 79    & 94 & 65 & 29 & 77    & 92 & 62 & 30   \\
                 & Optimal: 0.076  & 83    & 83 & 84 & 1  & 80    & 80 & 79 & 1    \\
                 \bottomrule  
\multicolumn{10}{p{420pt}}{ M. acc means macro accuracy. TNPR and TPR are true non-pushing rate and true pushing rate, respectively.}\\                 
\end{tabular}
\label{tab:EfficientNetB0evaluation}
\end{table*}

The performance results of the proposed framework, as well as the baselines, are presented in~\cref{tab:EfficientNetB0evaluation} and visualized in~\cref{fig:comparison-optimal-default}. The findings indicate that EfficientNetV1B0 with optimal threshold leads the framework to achieve superior macro accuracy and AUC with balanced TPR and TNPR compared to CNNs used in baselines 1-5. This can be attributed to the architecture of EfficientNetV1B0, which primarily relies on the Mobile Inverted Residual Bottleneck Convolution with relatively few parameters. As such, the architectural design proves to be particularly suited for the generated dataset focusing on local regions. The visualization in~\cref{fig:thresholds} shows the optimal threshold values for the baselines. These thresholds, as shown in~\cref{tab:EfficientNetB0evaluation} and~\cref{fig:comparison-optimal-default}, mostly improved the macro accuracy, TPR,  and balanced TPR and TNPR in the baselines. For example, baseline 1 with optimal threshold achieved \qty{84}{\percent} macro accuracy, roughly similar to the proposed framework. However,  it fell short of achieving a balanced TPR and TNPR along with improving TPR on both test sets as effectively as the framework. To provide further clarity, baseline 1 achieved  \qty{80}{\percent} TPR with  \qty{8}{\percent} as the difference between TPR and TNPR, whereas the proposed framework attained an \qty{86}{\percent} TPR with  \qty{2}{\percent} as the difference between TPR and TNPR on test set 1. Similarly, on test set 2, the framework achieved \qty{81}{\percent}  TPR, while baseline 1 achieved a TPR of \qty{74}{\percent}.

\begin{table*}[t]
\centering
\footnotesize
\caption{Performance results of the baseline 6.}
\begin{tabular}{lcccccc } 
\toprule
                                   & \multicolumn{3}{c}{Test set 1 \%} & \multicolumn{3}{c}{Test set 2 \%} \\   \cmidrule(r){2-4}   \cmidrule(r){5-7}
Threshold &
Macro accuracy  &
TNPR   &
TPR  &
Macro accuracy  &
TNPR   &
TPR  
  \\ \midrule
Default: 0.5          & 59    & 97      & 18    & 58    &59    & 57      \\    Optimal: 0.342 & 67    & 91    & 44         & 59    & 38    & 79 \\ \bottomrule 
\multicolumn{7}{p{340pt}}{TNPR and TPR are true non-pushing rate (Sensitivity) and true pushing rate (Specificity), respectively.}\\  
\end{tabular}
\label{tab:nodeepfeatureextraction}
\end{table*}

\begin{figure}[h]
   \centering
   \includegraphics[width=0.5\textwidth]{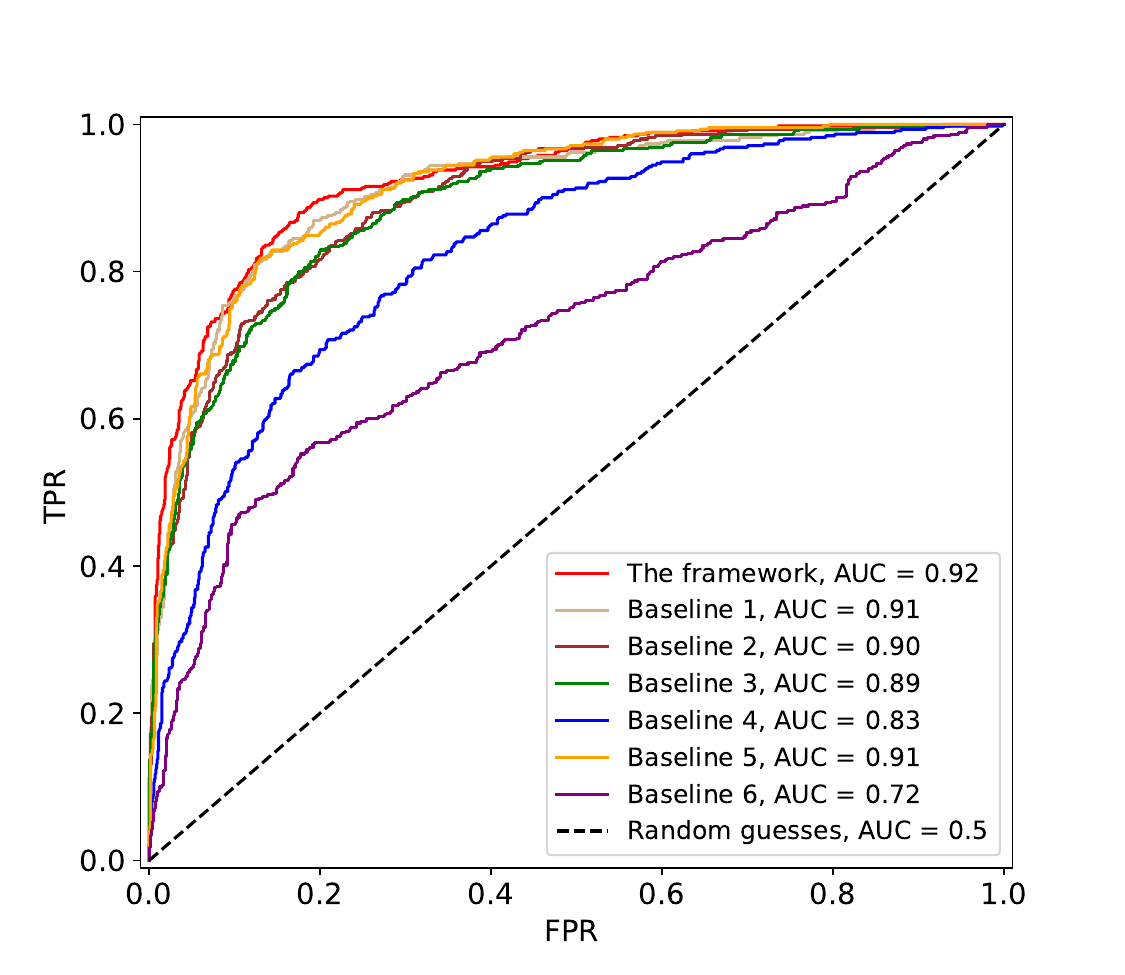}
  \caption{ROC curves with AUC values on the test set 1. Comparison between the introduced framework (based on EfficientNetV1B0) with five baselines based on different CNN architectures, as well as the one baseline without the deep feature extraction module (baseline 6). TPR stands for true pushing rate, while FPR refers to false pushing rate.  ROC represents Receiver Operating Characteristics. AUC stands for the area under the ROC Curve.}
  \label{fig:auc-roc-test1-all}
\end{figure}

\begin{figure}
   \centering
   \includegraphics[width=0.5\textwidth]{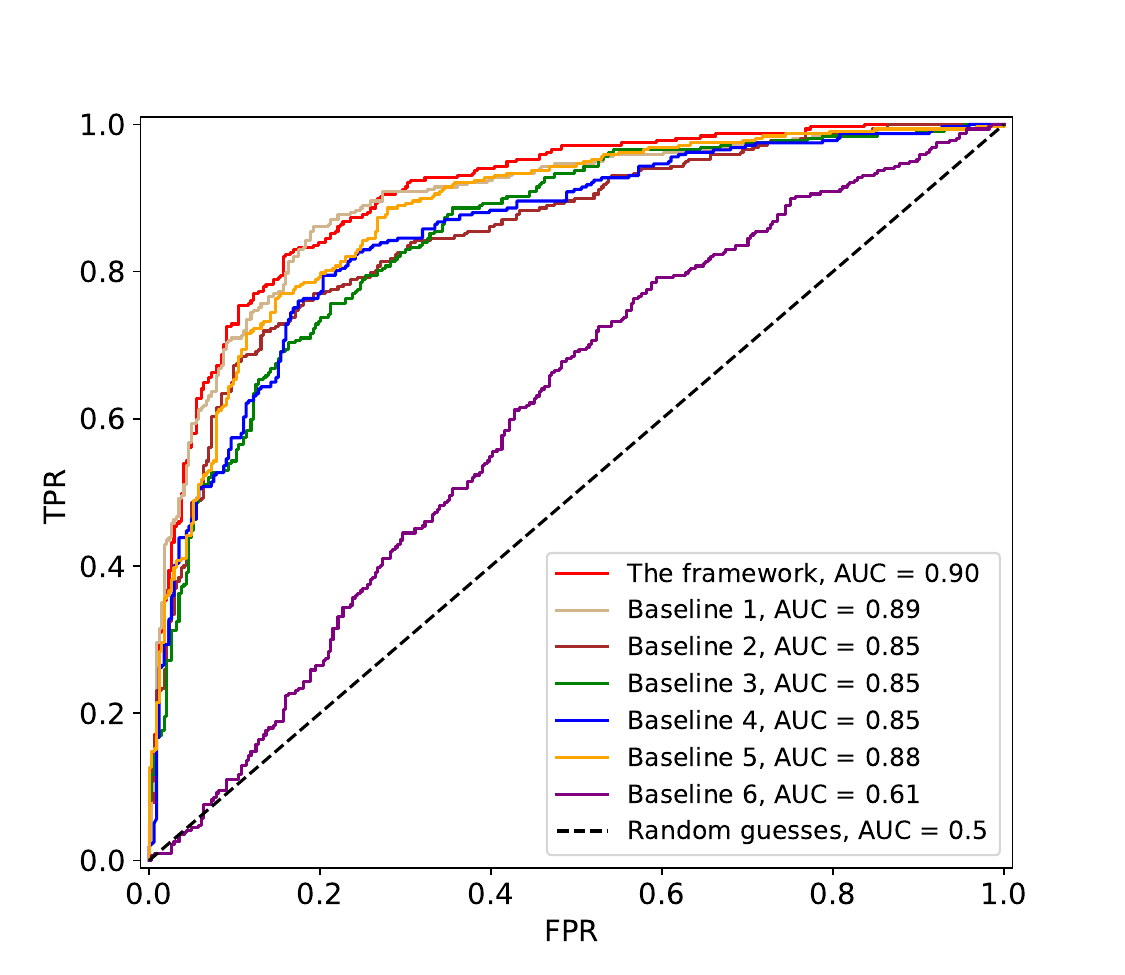}
  \caption{ROC curves with AUC values on the test set 2. Comparison between the framework (based on EfficientNetV1B0) with five baselines based on different CNN architectures, as well as the one baseline without the deep feature extraction module (baseline 6). TPR stands for true pushing rate, while FPR refers to false pushing rate.   ROC represents Receiver Operating Characteristics. AUC stands for the area under the ROC Curve.}
   
  \label{fig:auc-roc-test2-all}
\end{figure}

Compared to other baselines that utilize optimal thresholds on test set 1, the proposed framework outperformed them regarding macro accuracy, TPR, and TNPR. Similarly, on test set 2, the framework surpasses all baselines except for the ResNet50-based baseline (baseline 4). However, it is essential to note that this baseline only achieved better TNPR, whereas the introduced framework excels in macro accuracy and TPR. As a result, the framework emerges as the superior choice on test set 2. To alleviate any confusion in the comparison, \cref{fig:auc-roc-test1-all} shows the ROC curves with AUC values compared to its baselines on test set 1. Likewise,  \cref{fig:auc-roc-test2-all}  depicts the same for test set 2. The AUC values show that the proposed framework achieved better performance than the baselines on both test sets. Moreover, they substantiate that EfficientNetV1B0 is the most suitable CNN for extracting deep features from the generated local region samples.

In conclusion, the experiments demonstrate that the proposed framework, utilizing EfficientNetV1B0, achieved the highest performance compared to the baselines relying on other CNN architectures on both test sets. Furthermore, the optimal thresholds in the developed framework and the baselines resulted in a significant improvement in the performance across both test sets.

\subsection{Impact of Deep Feature Extraction Module}
 This section aims to investigate how the deep feature extraction module affects the framework's performance. For this purpose, a new baseline (baseline 6) is developed, incorporating a Voronoi-based local region extraction module and labeling component. In other words, the deep feature extraction module is removed from the proposed framework to construct this baseline.

\cref{tab:nodeepfeatureextraction} demonstrates that the baseline exhibited poor performance, with macro accuracy of \qty{67}{\percent} on test set 1 and \qty{59}{\percent} on test set 2. Additionally, \cref{fig:auc-roc-test1-all} and \cref{fig:auc-roc-test2-all} illustrate AUC values of \qty{72}{\percent} on test set 1 and \qty{61}{\percent} on test set 2 for baseline 6. Comparing this baseline with the weakest baseline in~\cref{tab:EfficientNetB0evaluation}, which utilizes ResNet50, it is evident that deep feature extraction leads to macro accuracy improvement of at least \qty{8}{\percent} on test set 1 and at least  \qty{20}{\percent} on test set 2. Similarly, deep feature extraction enhances AUC values by at least \qty{11}{\percent} on test set 1 and more than \qty{24}{\percent} on test set 2.

In summary, the deep feature extraction module significantly enhances the performance of the framework.

\subsection{Impact of Local Region Extraction}

The primary goal of this section is to evaluate the impact of the Voronoi-based local region extraction module on the performance of the proposed framework. To accomplish this, firstly, baseline 7 was created, which replaces this module with a new one that relies on static dimensions; to extract a local square region for each individual. In this new module, the target person’s position serves as the center of the extracted area, and each square region dimension is roughly 60 cm on the ground. Such dimension is enough to make the region contains the target person with his/her surrounding spaces. 

\begin{figure*}[b]
   \centering
   \includegraphics[width=1\textwidth]{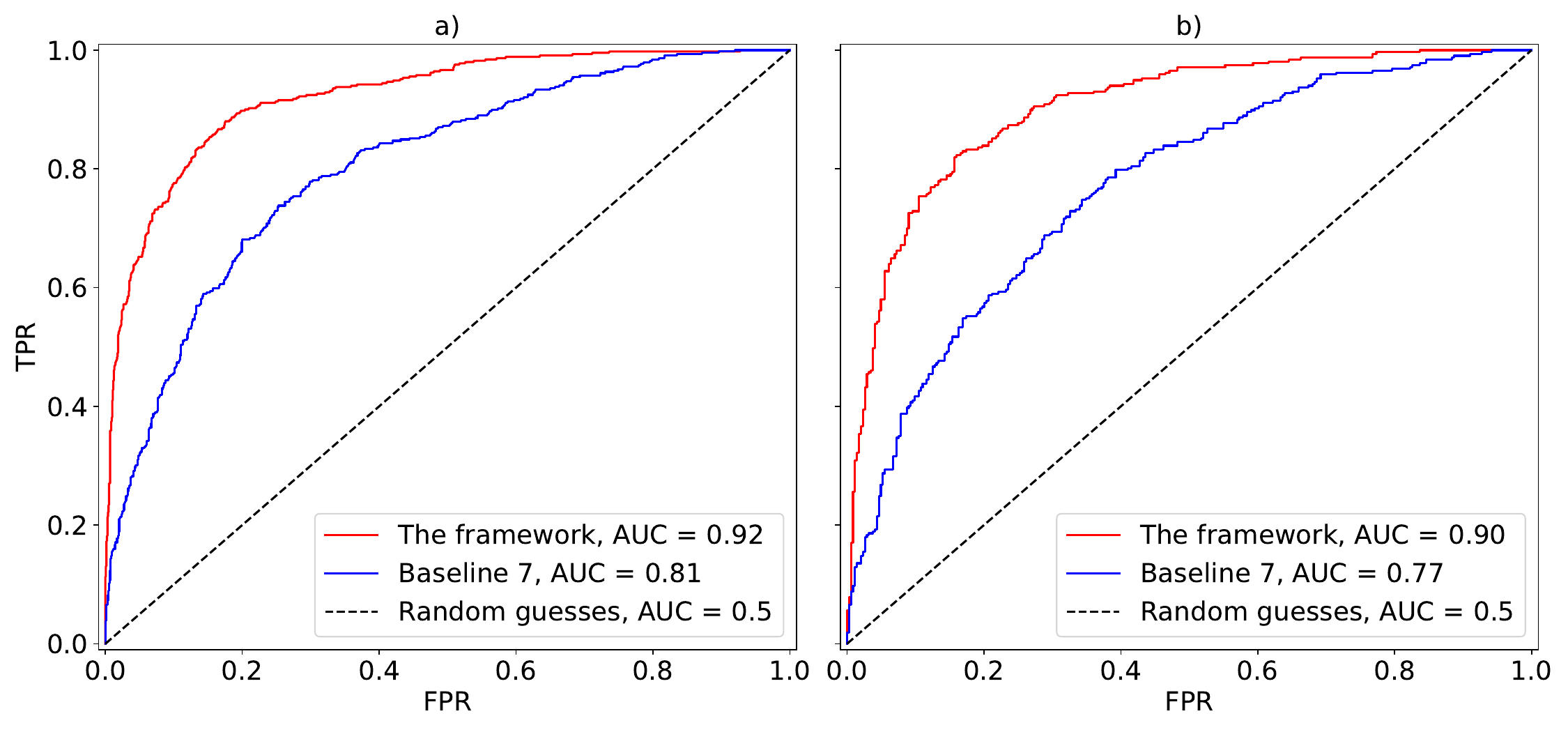}
  \caption{ROC curves with AUC values of the proposed framework against baseline 7 on a) test set 1 and b) test set 2. TPR stands for true pushing rate, while FPR refers to false pushing rate.  ROC represents Receiver Operating Characteristics. AUC stands for the area under the ROC Curve. }
  \label{fig:lrevaluation}
\end{figure*}

\begin{figure}[h]
   \centering
   \includegraphics[width=0.4\textwidth]{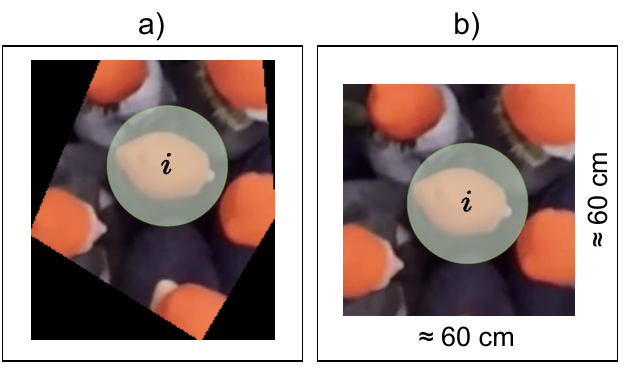}
  \caption{a) An example of a polygonal local region based on the bounded Voronoi Diagram. b) An example of a square local region based on static dimension. $i$ stands for the target person. }
  \label{fig:lr-static}
\end{figure}
\begin{table*}[t]
\centering
\footnotesize
\caption{Comparison to baseline 7.}
\begin{tabular}{lccccccc } 
\toprule
                               &    & \multicolumn{3}{c}{Test 1 set \%} & \multicolumn{3}{c}{Test 2 set \%} \\   \cmidrule(r){3-5}   \cmidrule(r){6-8}
Framework &
Optimal threshold &
Macro accuracy  &
TNPR   &
TPR  &
Macro accuracy  &
TNPR   &
TPR  
  \\ \midrule
The framework & 0.5         & 85   & 84      & 86    & 82    &81    & 83      \\ 
Baseline 7 & 0.241 & 79   & 78    & 82         & 62    & 42    & 82 \\ \bottomrule 
\multicolumn{7}{p{340pt}}{TNPR and TPR are true non-pushing rate (Sensitivity) and true pushing rate (Specificity), respectively.}\\  
\end{tabular}
\label{tab:staticsquare}
\end{table*}

\cref{fig:lr-static}b shows an example of a square local region of a target person ($i$). Then, a new dataset was generated utilizing the same video experiments and the same splitting technique used in preparing the local region dataset (\cref{tab:lableddataset}) to train and evaluate baseline 7. The main difference is that the samples in this new dataset are static square local regions (\cref{fig:lr-static}b) instead of dynamic polygonal regions (\cref{fig:lr-static}a).  According to the data presented in~\cref{tab:staticsquare}, baseline 7 achieved a macro accuracy of \qty{79}{\percent} on test set 1 and \qty{62}{\percent} on test set 2. This indicates that the Voronoi-based method results in \qty{6}{\percent} improvement in accuracy for test set 1 and a significant \qty{20}{\percent} improvement for test set 2. Additionally, \cref{fig:lrevaluation} demonstrates that the module enhanced the AUC value by \qty{11}{\percent} for test set 1 and \qty{13}{\percent} for test set 2.

In summary, the Voronoi-based local region extraction module enhanced the accuracy of the propsed framework by a minimum of \qty{6}{\percent}. This indicates that the Voronoi module is more effective than the new local region extraction in guiding the framework to identify relevant features from the input video.

\section{Conclusion and Future Work}
\label{sec:conclusionandfuturework}

This article introduced a new framework for automatically identifying pushing at the microscopic level within video recordings of crowds. The proposed framework utilizes a novel Voronoi-based method to determine the local region of each person in the input video over time. It further applies EfficientNetV1B0 to extract deep features from these local regions, capturing valuable information about individual behavior. Finally, a fully connected layer with a Sigmoid activation function is employed to analyze the deep features and annotate the pushing persons over time in the input video. To train and evaluate the performance of the framework, a novel dataset was created using six real-world experiments with their trajectory data and corresponding ground truths.The experimental findings demonstrated that the proposed framework surpassed seven baseline methods in terms of macro accuracy, true pushing rate, and true non-pushing rate.

The proposed framework has some limitations. First, it was designed to work exclusively with top-view camera video recordings that include trajectory data. Second, it was trained and evaluated based on a limited number of real-world experiments, which may impact its generalizability to a broader range of scenarios.
Our future goals include improving the framework in two key areas: 1) Enabling it to detect pushing persons from video recordings without the need for trajectory data as input. 2)  Improving its performance in terms of macro accuracy, true pushing rate, and true non-pushing rate by utilizing video recordings of additional real-world experiments and transfer learning techniques.

\bmhead{Acknowledgments}
The authors are thankful to Anna Sieben, Helena Lügering, and Ezel Üsten for the valuable discussions, manual annotation of the pushing behavior in the video of the experiments.

\section*{Declarations}

\textbf{Conflict of interest} The authors declare that there is no conflict of interests regarding the publication of this article.

\hfill \break
\noindent \textbf{Ethical approval}
The experiments used in the dataset were conducted according to the guidelines of the Declaration of Helsinki and approved by the ethics board at the University of Wuppertal, Germany. Informed consent was obtained from all subjects involved in the experiments.

\hfill \break
\noindent  \textbf{Funding} This work was funded by the German Federal Ministry of Education and Research (BMBF: funding number 01DH16027) within the Palestinian-German Science Bridge project framework, and partially by the Deutsche Forschungsgemeinschaft (DFG, German Research Foundation)—491111487.

\hfill \break
\noindent  \textbf{Availability of data and code} All videos and trajectory data used in generating the patch-based dataset were obtained from the data archive hosted by the Forschungszentrum Jülich under CC Attribution 4.0 International license~\cite{entrance2,crowdqueue}.  The implementation of the proposed framework, codes used for building training and evaluating the models, as well as  test sets and trained models are publicly available at: \url{https://github.com/PedestrianDynamics/VCNN4PuDe
} or at~\cite{alia_ahmed_2023_8175476} (accessed on 23 July 2023).   The training and validation sets are available from the corresponding authors upon request.

\hfill \break
\noindent  \textbf{Authors’ contributions}
Conceptualization, A.A.;
methodology, A.A., A.S.;
software, A.A.;
validation, A.A.;
formal analysis, A.A.;
investigation, A.A.;
data curation, A.A.;
writing---original draft preparation, A.A.;
writing---review and editing, A.A., M.M., M.C. and A.S.;
supervision, M.M., M.C. and A.S.;
 
 \noindent All authors have read and agreed to the published version of the~manuscript.
 
\hfill \break
\noindent \textbf{Open Access} This article is licensed under a Creative Commons
Attribution 4.0 International License, which permits use, sharing, adaptation distribution and reproduction in any medium or format, as long as you give appropriate credit to the original author(s) and the source, provide a link to the Creative Commons licence, and indicatecate if changes were made. The images or other third party material
in this article are included in the article’s Creative Commons licence,
unless indicated otherwise in a credit line to the material. If material is not included in the article’s Creative Commons licence and your intended use is not permitted by statutory regulation or exceeds the permitted use, you will need to obtain permission directly from the copyright holder. To view a copy of this licence, visit http://creativecommons.org/licenses/by/4.0/.

\bibliographystyle{unsrt}
\bibliography{references}

\end{document}